# Visual Reasoning and Multi-Agent Approach in Multimodal Large Language Models (MLLMs): Solving TSP and mTSP Combinatorial Challenges

Mohammed Elhenawy[1], Ahmad Abutahoun[2], Taqwa I.Alhadidi[2], Ahmed Jaber[3], Huthaifa I. Ashqar[4], Shadi Jaradat[1], Ahmed Abdelhay[5], Sebastien Glaser[1], and Andry Rakotonirainy[1*]

1 Accident Research and Road Safety Queensland, Queensland University of Technology, Brisbane, 130 Victoria Park Rd, Kelvin Grove QLD 4059, Australia
2 Civil Engineering Department, Al-Ahliyya Amman University, Amman, Jordan, 19328.
3 Department of Transport Technology and Economics, Faculty of Transportation Engineering and Vehicle Engineering, Budapest University of Technology and Economics, Műegyetem rkp. 3., H-1111 Budapest, Hungary.
4 Arab American University, Jenin, Palestine and Columbia University, NY, USA.
5 Computer and Systems Engineering Department, Faculty of Engineering, Minia University, Minia, Egypt
* Correspondence: r.andry@qut.edu.au

**Abstract:** Multimodal Large Language Models (MLLMs) harness comprehensive knowledge spanning text, images, and audio to adeptly tackle complex problems, including zero-shot in-context learning scenarios. This study explores the ability of MLLMs in visually solving the Traveling Salesman Problem (TSP) and Multiple Traveling Salesman Problem (mTSP) using images that portray point distributions on a two-dimensional plane. We introduce a novel approach employing multiple specialized agents within the MLLM framework, each dedicated to optimizing solutions for these combinatorial challenges. Our experimental investigation includes rigorous evaluations across zero-shot settings and introduces innovative multi-agent zero-shot in-context scenarios. The results demonstrated that both multi-agent models—Multi-Agent 1, which includes the Initializer, Critic, and Scorer agents, and Multi-Agent 2, which comprises only the Initializer and Critic agents—significantly improved solution quality for TSP and mTSP problems. Multi-Agent 1 excelled in environments requiring detailed route refinement and evaluation, providing a robust framework for sophisticated optimizations. In contrast, Multi-Agent 2, focusing on iterative refinements by the Initializer and Critic, proved effective for rapid decision-making scenarios. These experiments yield promising outcomes, showcasing the robust visual reasoning capabilities of MLLMs in addressing diverse combinatorial problems. The findings underscore the potential of MLLMs as powerful tools in computational optimization, offering insights that could inspire further advancements in this promising field. Project link: https://github.com/ahmed-abdulhuy/Solving-TSP-and-mTSP-Combinatorial-Challenges-using-Visual-Reasoning-and-Multi-Agent-Approach-MLLMs-.git

## 1. Introduction

The Traveling Salesman Problem (TSP) and its multi-salesmen variant (mTSP) are classic challenges in combinatorial optimization, recognized for their NP-hard complexity and practical relevance in logistics, planning, and network design [1,2]. TSP involves finding the shortest route for a single salesman to visit a set of cities and return to the start, while mTSP extends this to several salesmen, each covering different subsets of cities. These problems are computationally tough, especially as the number of cities grows, despite many advanced algorithms developed over the years.

Traditional solutions to TSP and mTSP typically rely on distance matrices and explicit calculations of routes based on node coordinates [3,4]. However, human problem-solving often employs visual and heuristic approaches to generate reasonable solutions without detailed calculations quickly. Inspired by these intuitive human strategies, our research explores a novel visual reasoning approach to solve TSP and mTSP. This approach leverages the power of visual inspection and team-based iterative refinement, bypassing the need for textual data or distance matrices, which are standard in computational methods [3,4].

In recent years, the application of meta-heuristics has become prevalent in addressing NP-hard combinatorial optimization problems such as the TSP [5]. These meta-heuristic approaches, like the Random-key Cuckoo Search, have shown promise not only in solving TSP but also in extending their applicability to other optimization problems like routing, scheduling, and mixed-integer programming [6]. Moreover, the integration of machine learning techniques has been explored to enhance the performance of existing solution algorithms for combinatorial optimization problems, including TSP [7]. Multimodal Large Language Models (MLLMs) have demonstrated proficiency in processing diverse modalities, including text, images, and audio, effectively addressing complex problems like the Traveling Salesman Problem (TSP) [44]. Researchers have investigated the potential of using Large Language Models (LLMs) like GPT-3.5 Turbo for solving the Traveling Salesman Problem (TSP), employing various approaches such as zero-shot in-context learning, few-shot in-context learning, and chain-of-thoughts (CoT) to improve solution quality [45]. Reinforced algorithms like the Lin-Kernighan-Helsgaun have demonstrated significant advancements in leveraging machine learning for solving TSP [8]. The TSP has been also extensively studied over decades using various heuristics such as iterated-local search, branch and bound, and graph coloring [9]. Researchers have also delved into parallelization strategies to optimize TSP solvers, with studies like the one by showcasing the capability of addressing large problem sizes efficiently [10]. Statistical analyses of frequency graphs related to TSP have been conducted to understand the problem better and devise effective resolution strategies within reasonable computation times [11]. Additionally, reinforcement learning algorithms have been compared for their performance in tackling TSP, highlighting the significance of exploring diverse approaches to address this challenging problem [12]. Various innovative methods have been proposed to tackle TSP, ranging from self-organizing map learning procedures to continuous relaxations and spider monkey optimization [13–15]. These approaches demonstrate the diversity of strategies employed to optimize TSP solutions, showcasing the interdisciplinary nature of research in this field. Furthermore, the utilization of linear function approximation and artificial immune system optimization reflects the continuous exploration of novel techniques to enhance combinatorial optimization outcomes [16,17].

The integration of different optimization algorithms like Ant Colony Optimization (ACO) and Genetic Algorithms (GA) has been a common theme in TSP research [18,19]. Studies have focused on improving these algorithms by incorporating local optimization heuristics and adaptive strategies to enhance their performance in solving TSP instances [20,21]. Moreover, the exploration of quantum-inspired methods and hybrid solvers has shown promising results in terms of computational efficiency and solution quality for TSP [22,23]. Researchers have also investigated the application of unconventional approaches such as affinity propagation clustering, producer-scrounger methods, and penguins search algorithms to optimize TSP solutions [15,24,25]. These diverse methodologies highlight the continuous quest for innovative solutions to complex combinatorial optimization problems like TSP. Additionally, the exploration of quantum annealers and hybrid solvers has opened new avenues for addressing TSP challenges, showcasing the evolving landscape of optimization techniques [23].

The advantages of using Multimodal Large Language Models (MLLMs) in solving problems include their ability to process and integrate diverse data types such as text [26,27], images [28,29], and videos [30], enabling a more holistic understanding of complex issues. The utilization of Multimodal Large Language Models (MLLM) in solving the TSP and mTSP is significant due to several compelling advantages. MLLMs can process and interpret visual data to suggest efficient routes through nodes, bypassing the need for textual information or distance matrices. This aligns with human visual problem-solving capabilities, allowing for a more intuitive and flexible approach to spatial problems. Moreover, MLLMs facilitate a collaborative, iterative process where multiple agents propose, evaluate, and refine solutions, leveraging diverse perspectives and expertise. This teamwork-based strategy not only enhances the quality of solutions by minimizing route intersections and optimizing lengths but also reduces the computational burden associated with traditional methods. Therefore, the use of MLLM in this context is promising as it showcases the potential of advanced AI to replicate human cognitive strategies, leading to significant improvements in solving complex problems like TSP and mTSP.

This study introduces two multi-agent strategies that utilize visual reasoning of MLLM to solve TSP and mTSP. The first strategy, labelled Multi-Agent 1, employs a trio of MLLM agents—Initializer, Critic, and Scorer—each with distinct roles in proposing, refining, and evaluating routes based on their visual quality. The second strategy, labelled Multi-Agent 2, simplifies the approach by using only the Initializer and Critic MLLM agents, focusing on rapid iterative refinement without the Scorer. Both methods aim to enhance route optimization by leveraging visual cues, mimicking the human ability to suggest and refine solutions without extensive computational resources. The study evaluates the efficacy of these visual reasoning-based strategies in solving the TSP and mTSP, comparing their performance against traditional zero-shot approaches using metrics such as mean gap percentage, standard deviation, and statistical validation through the Wilcoxon signed-rank test. Our results demonstrate significant improvements in solution quality and consistency, particularly in smaller and moderately sized problem instances, underscoring the potential of visually driven methodologies in complex problem-solving scenarios. The main contributions of this paper are as follows:

1. We introduce novel strategies for solving the TSP and mTSP using visual reasoning of MLLM alone, bypassing traditional numerical data like node coordinates or distance matrices.
2. We present MLLM as a multi-agent system—Initializer, Critic, and optionally Scorer agents—that iteratively refines routes. The Initializer suggests routes visually, the Critic improves them iteratively, and the Scorer evaluates them based on visual clarity and efficiency.
3. By leveraging iterative refinement, our approach minimizes route intersections, optimizes lengths, and ensures comprehensive node coverage without relying on numerical computations.

The remainder of this paper is organized as follows: In Section 2, we provided relevant work from the literature. In Section 3, we describe our approach and methods for in-context prompting. Section 4 presents the results of experimented methodologies, with a more extended discussion in Section 5. Finally, Section 6 concludes the study findings.

## 2. Related Work

Recent literature in multi-agent systems (MAS) has increasingly integrated deep reinforcement learning (DRL), as evidenced by Gronauer and Diepold [31]. This review outlines how DRL methods are structured to train multiple agents, emphasizing their applications in cooperative, competitive, and mixed scenarios. It also identified and addressed specific challenges unique to MAS, proposing strategies to overcome these obstacles and suggesting future research directions. Complementing this perspective, Dorri et al. [32] provided a comprehensive overview of MAS, discussing its definitions, features, applications, and challenges. Similarly, in Pop et al. [33], researchers evaluated various visual reasoning methodologies against traditional computational techniques. The study comprehensively surveyed mathematical formulations, solution approaches, and the latest advances regarding the generalized traveling salesman problem. Moreover, Yang et al. conducted alike survey on the cooperative control of multi-agent systems.

Currently, researchers are exploring the use of Large Language Models (LLMs) to address combinatorial problems like the traveling salesman problem. Liu et al.[34] introduced the concept of LLM-driven evolutionary algorithms (LMEA), marking the first attempt to apply LLMs in solving such problems. Meanwhile, Yang et al. [35] proposed Optimization by PROmpting (OPRO), where the LLM generates solutions from prompts containing previously generated solutions and their evaluations. This iterative process enhances solution quality with each step. Additionally, ensemble learning methods combined with LLMs, as discussed in works by Silviu et al. [36] have shown promising results in optimizing solutions.

In addressing the mTSP, Zheng et al. [37] focused on optimizing two objectives: minimizing the total length of all tours (minsum objective) and minimizing the length of the longest tour (minmax objective) among all salesmen. Furthermore, combining LLMs with other optimization techniques demonstrates potential for improving TSP solutions. Each method, whether utilizing zero-shot, few-shot, or Chain-of-thoughts (CoT) prompting techniques, aimed to enhance the accuracy of LLM responses, as evidenced in recent studies [38,39].

Additionally, a study of Bérczi et al. [40] considered a further generalization of mTSP, the many-visits mTSP, where each city has a request of how many times it should be visited by the salesmen. The authors provided polynomial-time algorithms for several variants of the many-visits mTSP that compute constant-factor approximate solutions. Huang et al. [41] explored LLMs' application in vehicle routing problems, demonstrating that direct input of natural language prompts enhances performance. They proposed a self-refinement framework to iteratively improve LLM-generated solutions, stressing the role of detailed task descriptions in boosting performance. Despite excelling in text-based tasks, LLMs face challenges with other data types [15]. However, MLLMs aim to overcome these limitations by integrating diverse data modalities (text, image, video, audio, etc.), broadening LLMs' potential applications beyond traditional text domains.

This study introduces two novel strategies for solving the TSP and mTSP that leverage the visual reasoning capabilities of MLLM. Unlike traditional approaches that rely on numerical data such as node coordinates or distance matrices, this study uses purely visual cues to infer efficient routes. This method mimics human visual problem-solving abilities and provides a more intuitive and flexible solution to spatial problems. While some recent research has explored the use of Large Language Models (LLMs) in combinatorial problems and optimization, such as LLM-driven evolutionary algorithms and Optimization by PROmpting (OPRO), these studies have primarily focused on text-based tasks. This study extends the application of LLMs to visual data, demonstrating the potential of MLLMs to address complex problem-solving scenarios without extensive computational resources.

Furthermore, this study fills the gaps identified in the existing literature by proposing a multi-agent system involving distinct roles for MLLM agents—Initializer, Critic, and Scorer. This collaborative, iterative approach enhances the quality of solutions through diverse perspectives and expertise, which is not extensively covered in prior research. While some studies have discussed the integration of DRL in multi-agent systems and the cooperative control of such systems, this study uniquely applies visual reasoning in a multi-agent context to optimize routes for TSP and mTSP. By focusing on visual reasoning and iterative refinement in MLLMs, the study minimizes route intersections, optimizes route lengths, and ensures comprehensive node coverage, thus offering a novel methodology that advances the current state of research in visual computational techniques and complex problem-solving.

## 3. Materials and Methods

Our methodology draws on two concepts inspired by human problem-solving approaches. The first is based on the human ability to suggest efficient routes through nodes without explicit calculations visually. The second concept leverages the advantages of teamwork, where individuals collaboratively refine solutions. In this approach, a team member proposes one or more solutions, which are collectively analyzed. Ineffective solutions are discarded while promising ones are iteratively improved through subsequent proposal, evaluation, and refinement cycles. This iterative team-based strategy enhances the development of practical solutions by leveraging diverse perspectives and expertise.

In this paper, we adopt these ideas to develop two strategies to solve the TSP and mTSP using visual reasoning alone, without relying on textual information about node locations or distance matrices with the use of MLLMs. The use of MLLMs in solving the TSP and mTSP is important because it harnesses the power of advanced AI to integrate visual reasoning and collaborative problem-solving. MLLM can process and interpret visual information, enabling the identification of efficient routes through nodes based solely on visual cues, without requiring textual data or distance matrices. This capability mirrors human visual problem-solving and enhances flexibility in dealing with complex spatial problems. Additionally, MLLM can facilitate iterative teamwork by analyzing and refining proposed solutions collaboratively, leveraging diverse inputs and perspectives to optimize outcomes. This approach not only improves the practical application of visual computational techniques but also demonstrates the potential for MLLM to revolutionize complex problem-solving by mimicking human cognitive processes. The following subsection will detail the proposed strategies, explaining how each leverages purely visual cues to infer efficient routes across multiple nodes, thus enhancing our understanding of visual computational capabilities in complex problem-solving.

### *3.1 Multi-Agent 1*

The proposed methodology in Fig. 1 adopts a multi-agent strategy to solve the m-salesmen problem where $m \in \{1,2,3\}$. The process involves three agents: the Initializer, the Critic, and the Scorer. Initializer Agent visually inspects the nodes' layout without textual information about their coordinates. Using visual reasoning, it suggests an initial order for visiting the nodes, aiming to propose a good solution. The suggested solution by the Initializer is visualized and presented to the Critic Agent along with a prompt instructing it to inspect the visualization and suggest better routes. The Critic Agent employs a self-ensemble method by raising its temperature to 0.7, generating seven solutions. These solutions are visualized and passed on to the Scorer Agent. This agent assigns a score to each solution based solely on its visual quality without calculating the actual distances. The solution with the highest score is then selected and returned to the Critic.

The steps involving the Critic and the Scorer continue iteratively until the maximum number of iterations is reached. Finally, the best solution is returned as the proposed solution to the m-salesmen problem. This strategy leverages a visually driven multi-agent strategy where the Initializer proposes, the Critic refines, and the Scorer evaluates solutions based on visual reasoning using MLLM to iteratively enhance the solution quality for the m-salesmen problem. In this strategy, we used ChatGPT-4o as the Initializer, the Critic, and the Scorer.

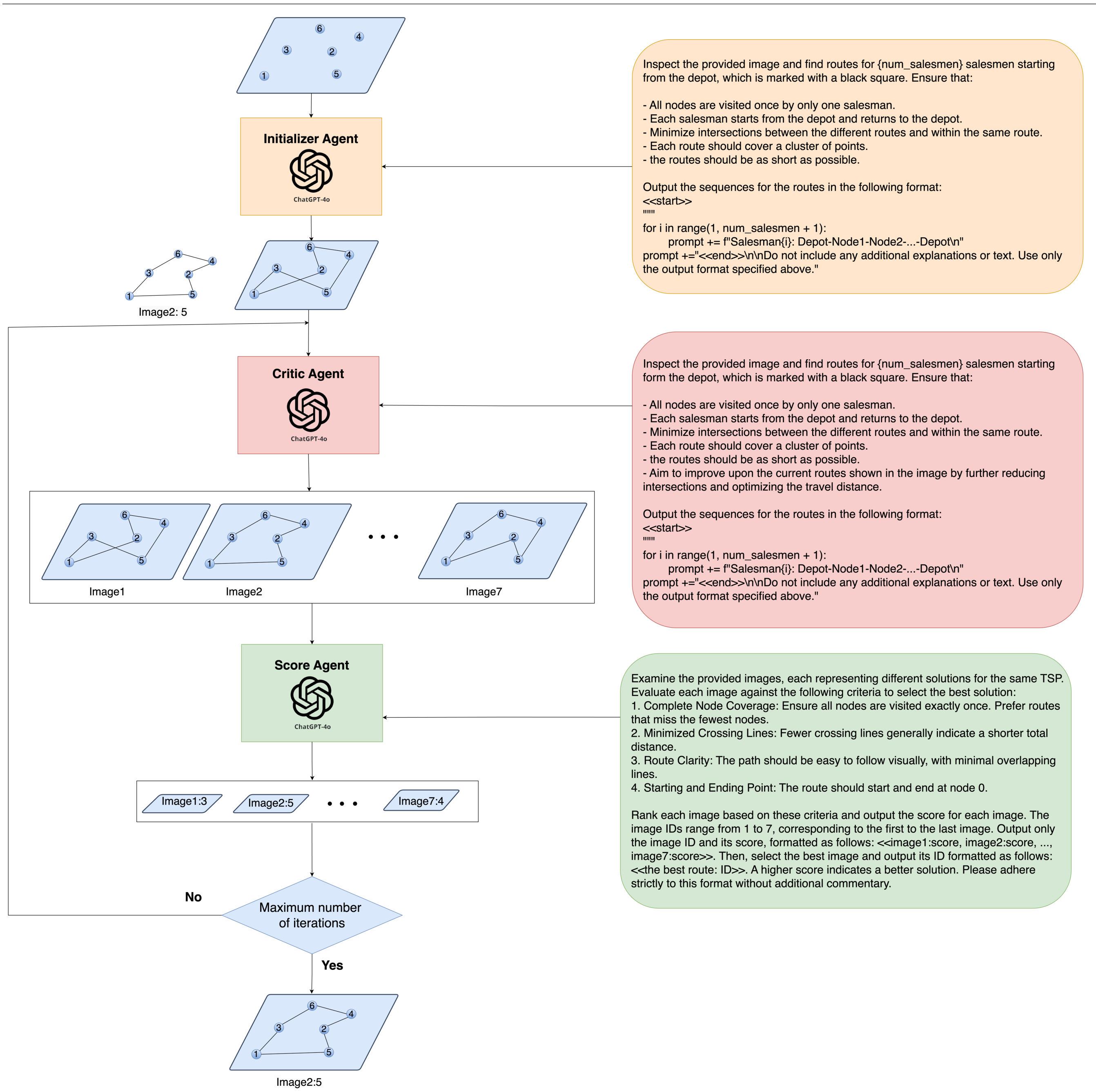

**Fig 1.** Multi-Agent 1 Strategy: The diagram illustrates a three-phase approach using visual reasoning, where an Initializer Agent proposes routes, a Critic Agent refines them, and a Scorer Agent evaluates their visual quality, enhancing the solutions iteratively.

### *3.2 Multi-Agent 2*

The simplified Multi-Agent 2 strategy, designed for solving the m-salesmen problem with fewer agents, uses only the Initializer and Critic agents, eliminating the Scorer agent. In this lighter approach, the Initializer Agent continues by visually inspecting the nodes and proposing an initial sequence for visiting them, leveraging visual cues instead of relying on coordinate data. This proposed route is then passed to the Critic Agent, which, unlike its role in Multi-Agent 1, operates under a higher temperature setting of 0.7 to generate a single alternative route per iteration. Each new solution is visually presented back to the Critic for further refinement. The iterative process between the Initializer and Critic contin-ues until the maximum number of iterations is reached, at which point the process halts. This streamlined version focuses on rapid iteration and refinement without the evaluative step of scoring, aiming to efficiently enhance route optimization through continuous visual feedback and critical adjustments. In this strategy, we used ChatGPT-4o as the Initializer and the Critic.

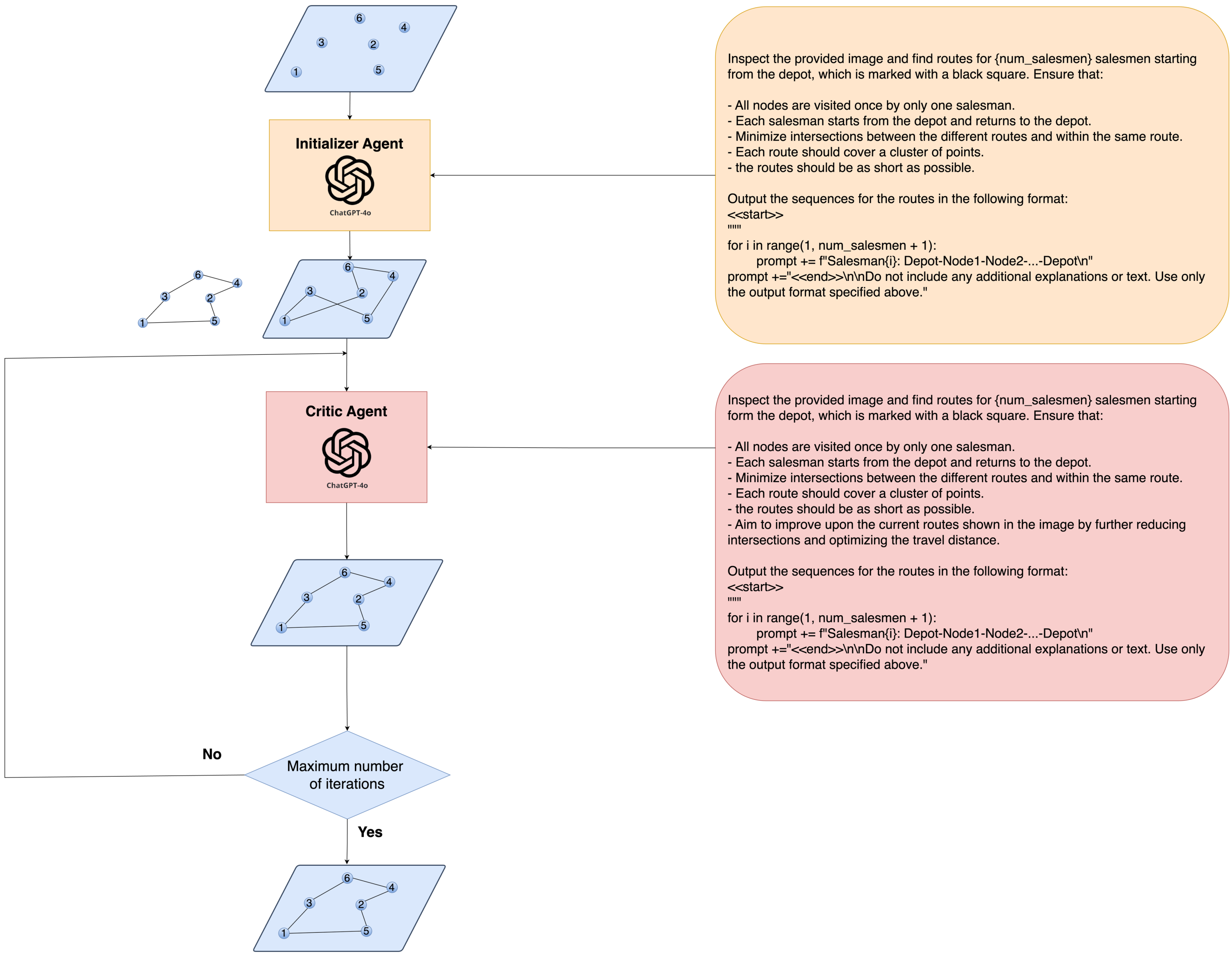

**Fig 2.** Multi-Agent 2 Strategy: The diagram illustrates a streamlined two-agent methodology, using visual reasoning without computational distance measures. The Initializer agent proposes routes, which the Critic agent refines, iterating until a maximum iteration

### *3.3 Prompts Engineering*

In this subsection, we detail the specific prompts used in ChatGPT-4o to direct the actions of the different agents involved in our multi-agent system for solving the m-salesmen problem using visual reasoning. Each agent is tasked with unique functions that contribute to identifying good routes based on visual inputs alone without relying on numerical data like distances or coordinates.

**Table 1.** Rules and Prompts

| Rule | Prompt |
| --- | --- |
| Initializer agent | f"""" Inspect the provided image and find routes for {num_salesmen} salesmen starting from the depot, which is marked with a black square. Ensure that: |
| | - All nodes are visited once by only one salesman.<br>- Each salesman starts from the depot and returns to the depot.<br>- Minimize intersections between the different routes and within the same route.<br>- Each route should cover a cluster of points.<br>- The routes should be as short as possible.<br>Output the sequences for the routes in the following format:<br><<start>><br>"""<br>for i in range(1, num_salesmen + 1):<br>    prompt += f"Salesman{i}: Depot-Node1-Node2-...-Depot\n"<br>prompt += "<<end>>\n\nDo not include any additional explanations or text. Use only the output format specified above." |
| Critic agent | f"""<br>Inspect the provided image and find routes for {num_salesmen} salesmen starting from the depot, which is marked with a black square. Ensure that:<br>- All nodes are visited once by only one salesman.<br>- Each salesman starts from the depot and returns to the depot.<br>- Minimize intersections between the different routes and within the same route.<br>- Each route should cover a cluster of points.<br>- The routes should be as short as possible.<br>- Aim to improve upon the current routes shown in the image by further reducing intersections and optimizing the travel distance.<br><br>Output the sequences for the routes in the following format:<br><<start>><br>"""<br>for i in range(1, num_salesmen + 1):<br>    prompt += f"Salesman{i}: Depot-Node1-Node2-...-Depot\n"<br>prompt += "<<end>>\n\nDo not include any additional explanations or text. Use only the output format specified above." |
| Score agent | Examine the provided images, each representing different solutions for the same TSP. Evaluate each image against the following criteria to select the best solution:<br>1. Complete Node Coverage: Ensure all nodes are visited exactly once. Prefer routes that miss the fewest nodes.<br>2. Minimized Crossing Lines: Fewer crossing lines generally indicate a shorter total distance.<br>3. Route Clarity: The path should be easy to follow visually, with minimal overlapping lines.<br>4. Starting and Ending Point: The route should start and end at node 0.<br>Rank each image based on these criteria and output the score for each image. The image IDs range from 1 to 7, corresponding to the first to the last image. Output only the image ID and its score, |
| | formatted as follows: <<image1: score, image2: score, …, image7: score>>. Then, select the best image and output its ID formatted as follows: <<the best route: ID>>. A higher score indicates a better solution. Please adhere strictly to this format without additional commentary. |

*3.4 Initializer Agent Prompt*

The prompt provided for the Initializer agent is designed to guide the agent in constructing routes for multiple salesmen (the exact number specified by {num_salesmen}) from a visual representation of nodes on an image. The prompt directs the agent as follows:

- Starting Point: The prompt specifies that the routes must start and end at a depot, marked as a black square in the image. This sets a clear starting and returning point for each salesman's route.
- Node Visitation: It's stipulated that all nodes must be visited exactly once by only one salesman, ensuring that the task covers all designated points without overlap between salesmen.
- Route Efficiency: The prompt demands minimizing intersections within and between routes. This aims to reduce potential route conflicts and ensures that the paths taken are as efficient as possible.
- Cluster Coverage: Each route should cover a cluster of points, implying that the agent should look for logical groupings of close nodes to form each route, enhancing practicality and efficiency.
- Route Length: The emphasis is also on keeping routes short, prioritizing direct paths and proximity among nodes within a route.
- Output Format: The expected output format is very structured, asking for the route of each salesman to be listed sequentially from the depot, through each node, and back to the depot. The format is specified to begin with <<start>> and end with <<end>>, and only the routes are to be listed without any additional text or explanation.

*3.5 Critic Agent Prompt*

The prompt for the Critic agent builds on the Initializer's task by refining the suggested routes. It is structured to guide the Critic in analyzing the existing routes provided by the Initializer and then improving them based on specific criteria. Here's how the prompt directs the agent:

- Initial Instructions: Like the Initializer agent, the Critic starts with the same basic guidelines regarding the depot, unique node visits by each salesman, minimizing route intersections, and covering clustered points.
- Optimization Focus: The key addition for the Critic is the instruction to "improve upon the current routes shown in the image by further reducing intersections and optimizing the travel distance." This pushes the agent to look for enhancements over the already suggested routes, focusing on increased efficiency and reduced travel distances.
- Output Format: The format for output remains structured and specific. The Critic must list each salesman's route starting and ending at the depot in a precise sequence without additional commentary. The sequence is strictly formatted from the depot to the last node and back, maintaining clarity and consistency.

*3.6 Score Agent Prompt*

The prompt for the Score agent is designed to evaluate and rank multiple solutions for the Traveling Salesman Problem (TSP) based on visual clarity and efficiency. Here's how the prompt directs the agent:

- Evaluation Criteria:
  - Complete Node Coverage: This ensures all nodes are visited precisely once, emphasizing routes that don't skip any nodes.
  - Minimized Crossing Lines: Routes with fewer intersections are preferred as they generally suggest a shorter overall path.

- Route Clarity: The paths should be straightforward to trace visually, with minimal overlapping, which enhances the readability and practicality of the route.
  - Starting and Ending Point: It's essential that the route begins and ends at the same point, labeled as node 0, ensuring a closed loop that is typical for TSP solutions.
- Scoring and Ranking: Each route is assigned a score based on how well it meets the above criteria. The scores are directly related to the route's efficiency and clarity. The prompt specifies that each image representing a different route solution should be scored and then listed with its corresponding score, maintaining a clear format for easy comparison.
- Output: The scores are to be formatted concisely: <<image1: score, image2: score, …, image7: score>>. Additionally, the highest-scoring route is to be highlighted as the best solution with its specific ID in the format: <<the best route: ID>>.
- Format and Procedure: The agent is instructed to strictly adhere to the output format without adding any supplementary explanations or textual content. This structured approach ensures that the outputs are standardized and focused solely on the numerical scoring and ranking.

### *3.7 Test Data*

The test data described in this paper is generated through a process where each node's coordinates ($x$ and $y$) are independently sampled from a uniform distribution ranging from zero to five. This method ensures that all data points are evenly and randomly distributed within a $5x5$ square area. The uniform distribution is used to simulate an equal probability of any point within the chosen specified range, which helps create a diverse set of scenarios for evaluating the model's performance on spatial problems like the TSP. This approach is typical in computational experiments where the goal is to assess algorithmic efficiency under varied yet controlled conditions to ensure that the performance metrics are not biased by any specific configuration of the nodes [4].

### *3.8 Ground Truth Solutions*

In this study, we employed the Google OR-Tools framework to address the mTSP [42,43], a variant of the TSP, where more than one salesman is involved. The mTSP is known for its NP-hard complexity, making exact solutions impractical for large datasets. To manage the routing challenges efficiently, we utilized the pywrapcp package . Routing Model from the OR-Tools suite, a powerful tool designed to streamline the process of defining, solving, and analyzing routing problems. The Routing Model is configured with a Euclidean distance matrix representing the distances between nodes. This matrix is crucial for the algorithm as it serves as the foundation upon which the routing decisions are based. To generate initial feasible solutions rapidly, we employed the SAVINGS heuristic, a commonly used approach for vehicle routing problems that provides a good starting solution for further improvement. This heuristic is integrated into the OR-Tools through the routing_enums_pb2.FirstSolutionStrategy.SAVINGS.

Further refinement of the initial solution is achieved using the GUIDED_LOCAL_SEARCH metaheuristic, which is part of the OR-Tools' local search algorithms. This metaheuristic helps navigate the solution space effectively, improving the quality of solutions by escaping local optima—a frequent challenge in route optimization problems. The local search parameters are tuned to balance between computation time and solution quality, making the approach feasible for larger datasets.

Although the solutions derived from this methodology are not guaranteed to be globally optimal due to the heuristic nature of the algorithms, they are generally close to optimal and computationally efficient [42,43]. This approach allows us to handle more significant problem instances that are typically infeasible with exact methods, providing practical solutions within a reasonable timeframe. Thus, the use of Google OR-Tools, particularly the pywrapcp and RoutingModel, represents a robust method for tackling complex routing problems like the mTSP in applied research settings [42,43].

## 4. Results

This section will explore the solution quality of the two proposed strategies.

*4.1 Multi-Agent 1 Solution Quality*

The presented analysis in Fig. 3 across three distinct scenarios—solving the traveling salesman problem with one, two, and three salesmen—demonstrates a comparative evaluation of solution quality measured by the mean and standard deviation of gap percentages as problem size increases. For configurations involving two and three salesmen, the observed negative mean gap values at smaller problem sizes indicate superior performance of the proposed strategies over the solutions generated by the Google OR tool within a fixed computation time of 120 seconds. This suggests that as the complexity introduced by additional salesmen increases, the proposed strategies are better equipped to handle such complexities, outperforming the baseline set by Google's optimization tools, particularly at smaller problem scales.

Further, the standard deviation of the gaps consistently illustrates lower variability for the Multi-Agent Strategy 1 compared to the zero-shot method across all problem sizes, indicating a more consistent output from the proposed strategies. This trend is consistent even as the problem size increases, underlining the challenges posed by larger problems and showcasing the proposed solutions' robustness.

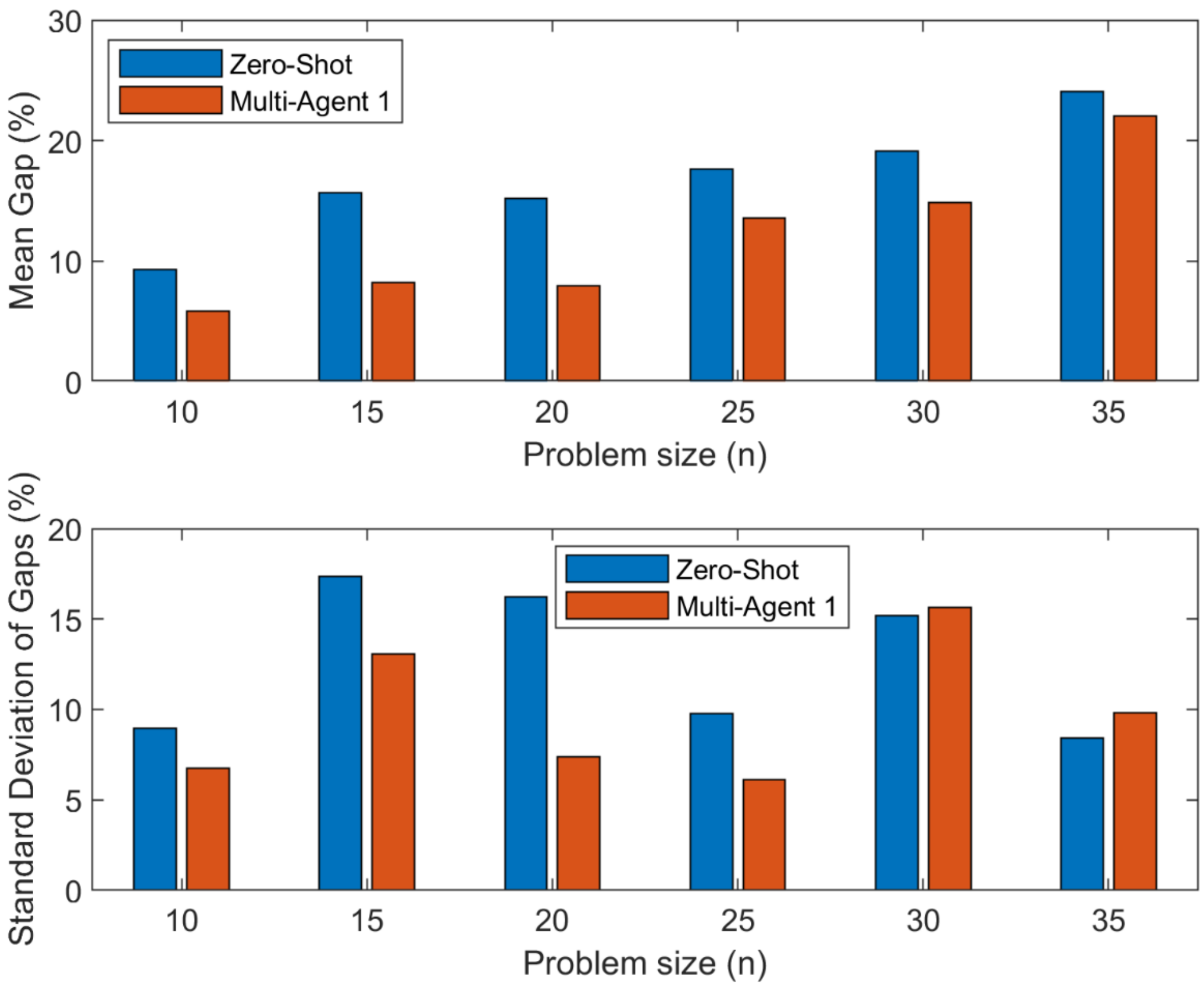

(a)

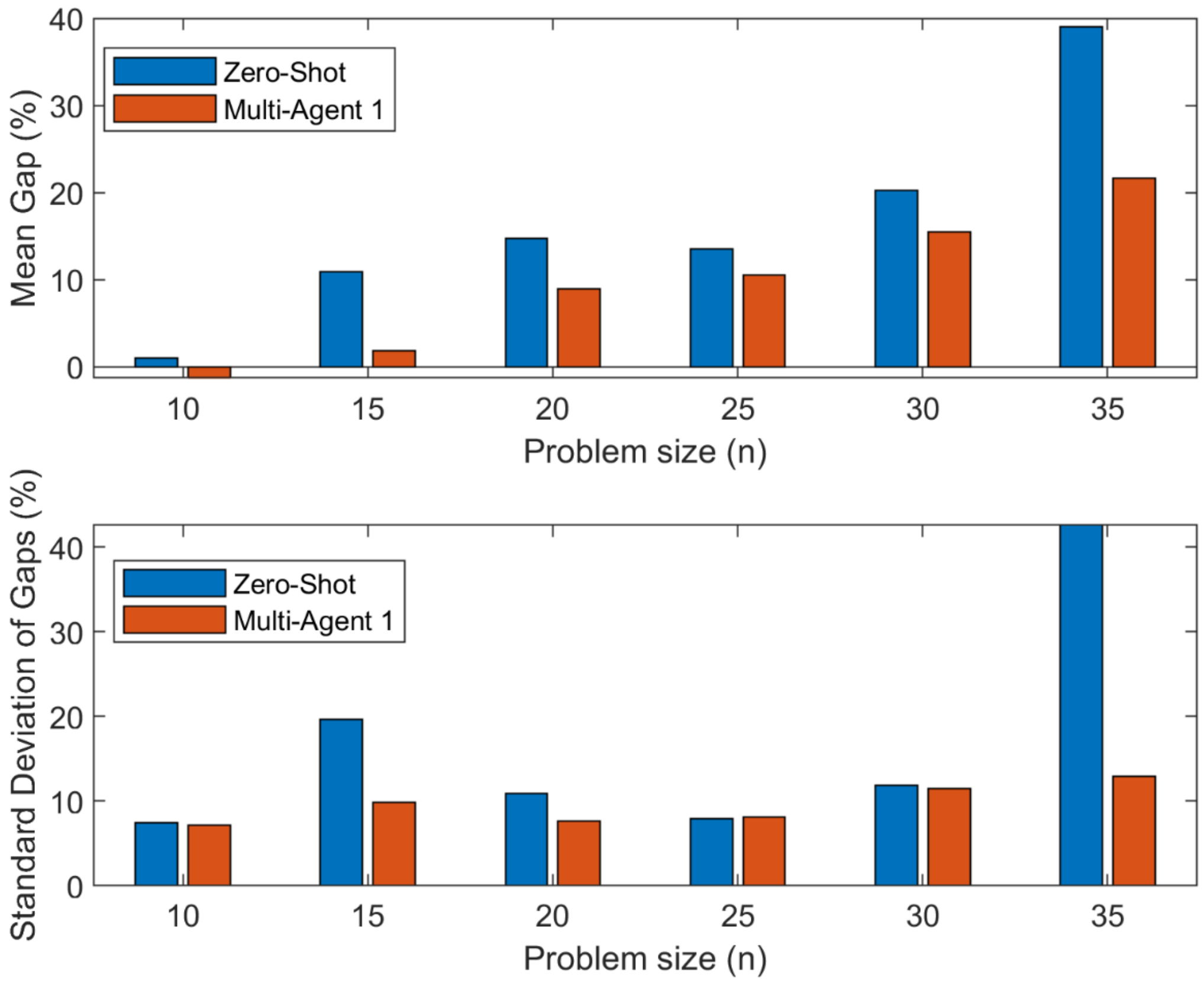

(b)

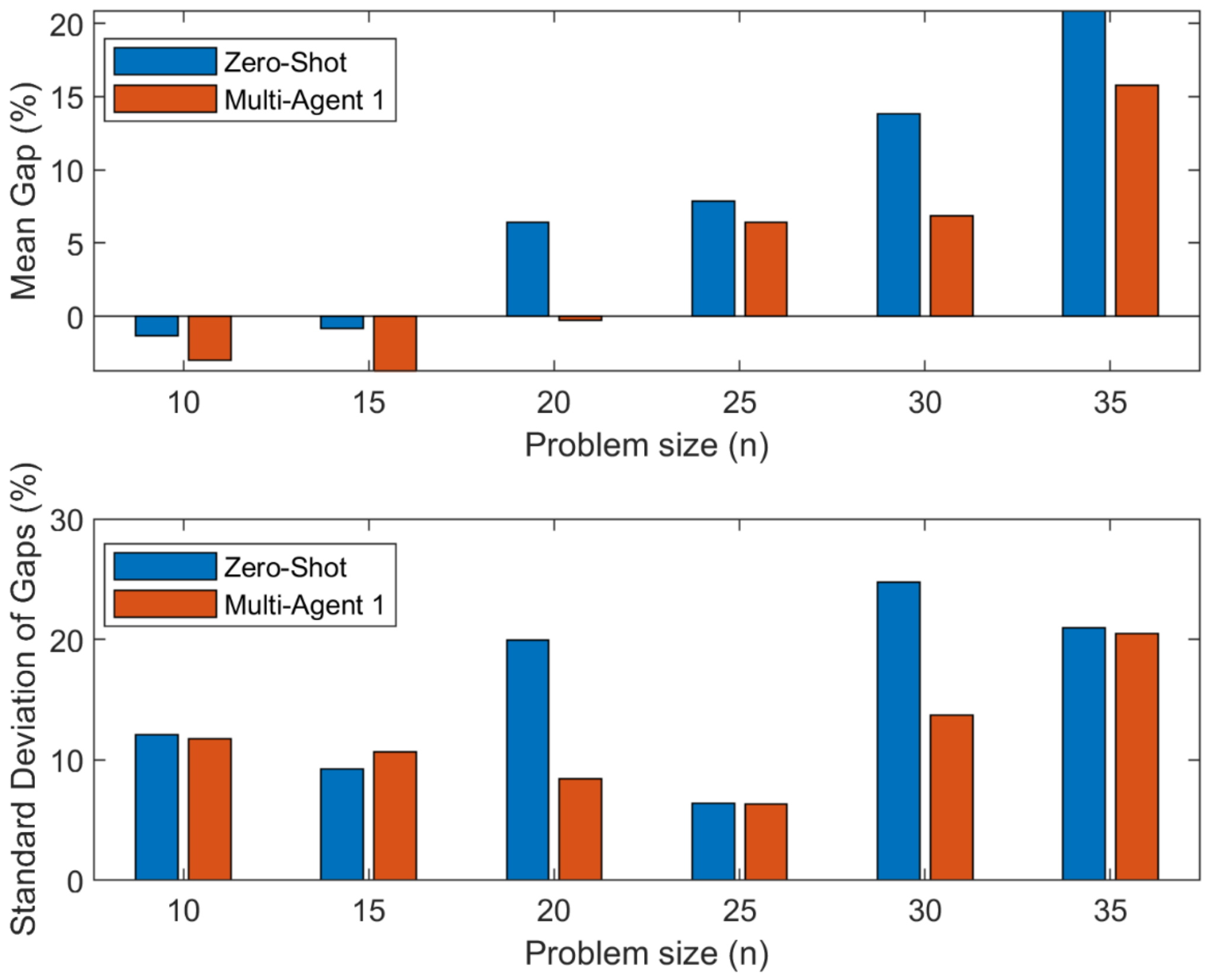

(c)

**Fig 3.** Comparative Performance of Zero-Shot and Multi-Agent Strategy 1 across Different Salesmen Configurations: Analysis of Mean Gap and Standard Deviation by Problem Size (a) one salesman (b) two salesmen (c) three salesmen.

Table 2 presents a comprehensive analysis using the Wilcoxon signed-rank test to statistically validate the improvement in route quality offered by the Multi-Agent 1 strategy over the Zero-Shot approach. In this paired, two-sided test, the null hypothesis posits that the difference between the routes generated by Zero-Shot and Multi-Agent 1 strategies (denoted as x - y) has a median of zero. The results indicate a significant improvement with the Multi-Agent 1 approach, as evidenced by the consistently low p-values across various problem sizes, particularly for smaller problems.

**Table 2.** Statistical Analysis of Route Quality Improvement Using Multi-Agent 1 Strategy Over Zero-Shot.

| | $m = 1$ | | $m = 2$ | | $m = 3$ | |
|---|---|---|---|---|---|---|
| problem size | p-value | number of pairs | p-value | number of pairs | p-value | number of pairs |
| 10 | 0.0010 | 27 | 0.0002 | 24 | 0.0004 | 29 |
| 15 | 0.0003 | 27 | 0.0001 | 28 | 0.0020 | 22 |
| 20 | 0.0004 | 24 | 0.0002 | 20 | 0.0010 | 22 |
| 25 | 0.0156 | 28 | 0.0005 | 18 | 0.0156 | 19 |
| 30 | 0.0020 | 16 | 0.0156 | 20 | 0.1250 | 10 |
| 35 | 0.5000 | 9 | 1.0000 | 7 | 0.0313 | 13 |

As problem sizes increase, the initializer agent encounters more difficulties in effectively managing node visits, leading to an increased rate of hallucinations where the proposed routes fail to visit all designated nodes. This effect of hallucinations decreases the number of valid problem instances available for paired comparisons. Hallucinations are more pronounced in larger problems due to the heightened complexity involved in routing.

The critic agent plays a pivotal role in this context by identifying and rectifying incomplete or erroneous routes that miss crucial nodes, thereby ensuring the routes are optimized to include all necessary stops. This intervention by the critic agent is essential for enhancing route quality, particularly when initial outputs from the initializer may be flawed. For our statistical analysis, we exclude any problem instances where the zero-shot strategy results in incomplete routes, reducing the number of pairs available for the Wilcoxon signed-rank test. This limitation can affect our ability to detect significant differences between the multi-agent 1 strategy and the zero-shot approach, especially when employing non-parametric tests that is less power than the parametric tests

### *4.2 Multi-Agent 1 Example 1*

In this example, the multi-agent system comprises an initializer and a critic agent and shows an illustrate of an incomplete route hallucination. The process starts with the initializer generating an initial solution which is visualized in the first row of the images. Notably, this initial solution misses one crucial node. The incomplete solution is then presented to the critic agent, which evaluates it and suggests seven possible solutions to improve the initial output. These solutions are visually represented in Fig. 4. Subsequently, these images are passed to a score agent, which assigns a score to each solution based on its effectiveness and completeness.

The scoring details reveal that solutions number 1 and 3 are identical, and both received the same score, which indicates consistency in the scoring process by the score agent. Furthermore, solutions number 4 and 7 achieved the highest score of 4, suggesting they are the most promising solutions offered by the critic. Following this scoring process, the solution with the highest score, in this case, solution number 4, is selected to be passed back to the critic agent for the next iteration. This iterative process aims at refining the solution until it covers all required nodes, which, as noted, happens after the first iteration—indicating that the initial solution's shortcomings are effectively addressed by the multi-agent system. This example illustrates the iterative interaction between the initializer, critic agent, and score agent in improving a solution within a multi-agent system, ultimately leading to a good solution.

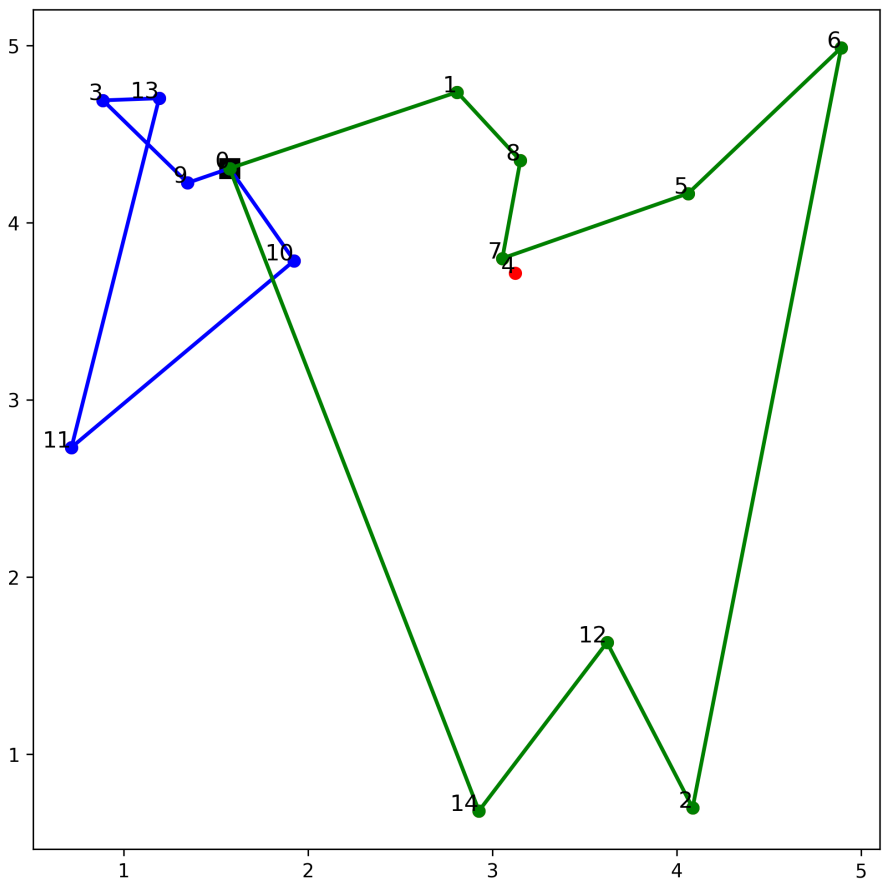

Initial solution

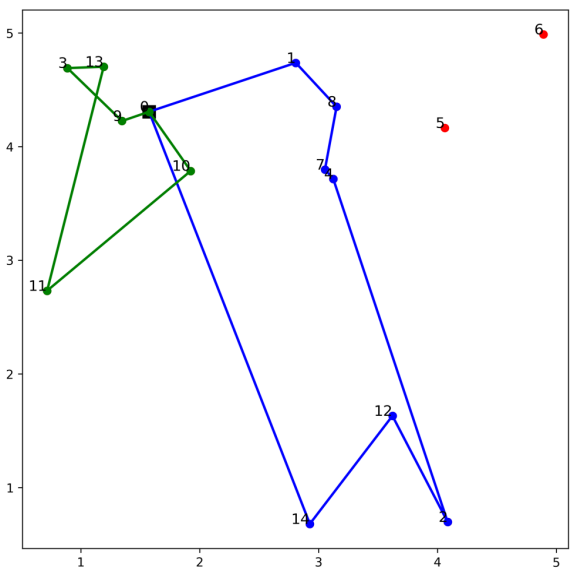

Suggested solution #1 score =3

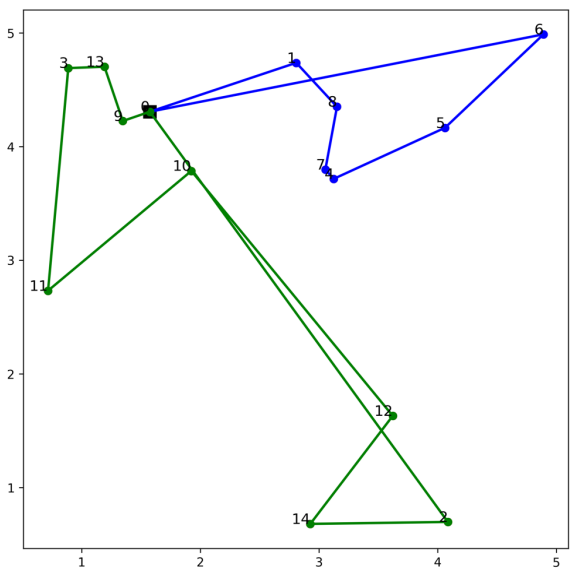

Suggested solution #2 score =2

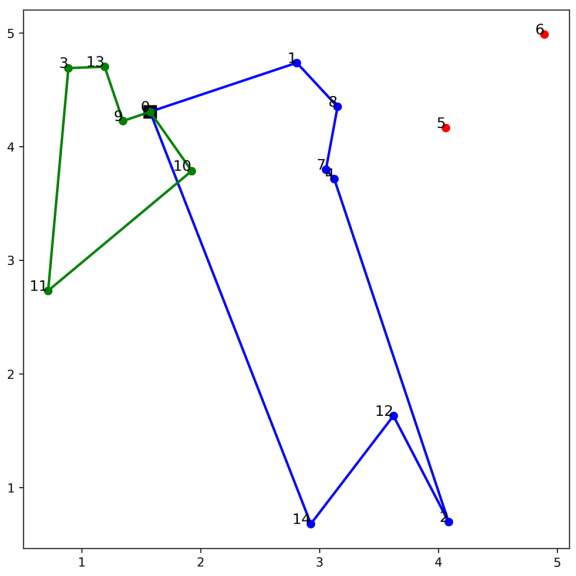

Suggested solution #3 score =3

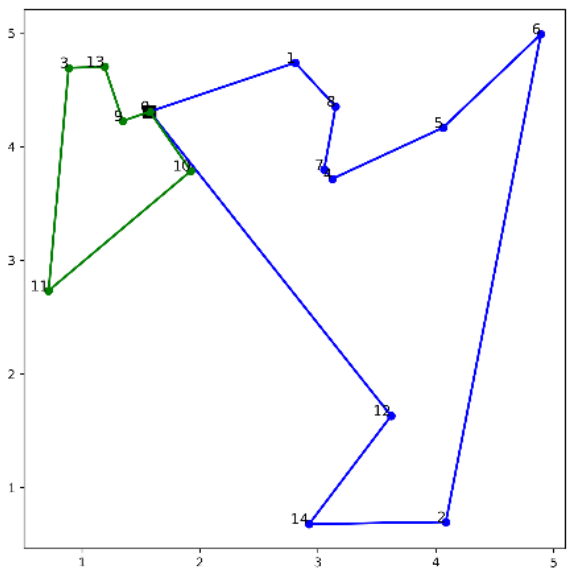

Suggested solution #4 score =4

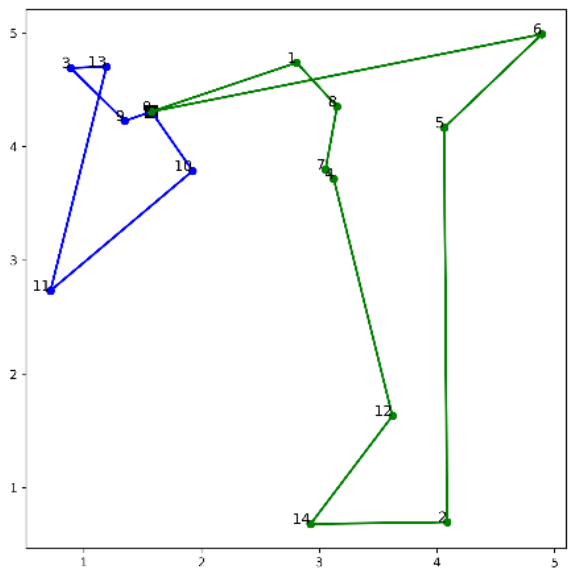

Suggested solution #5 score =1

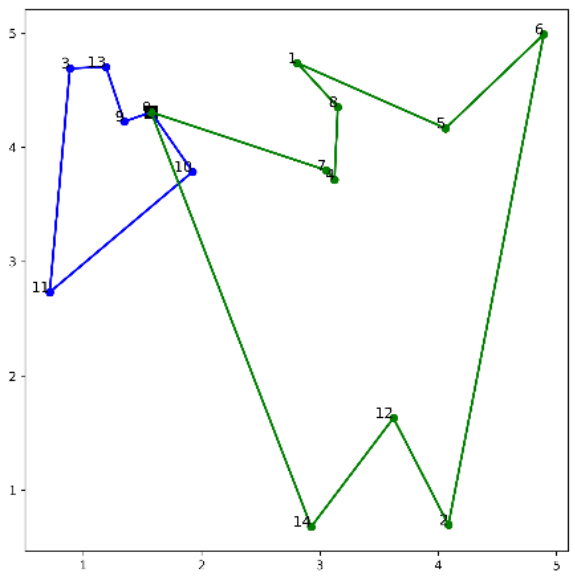

Suggested solution #6 score =2

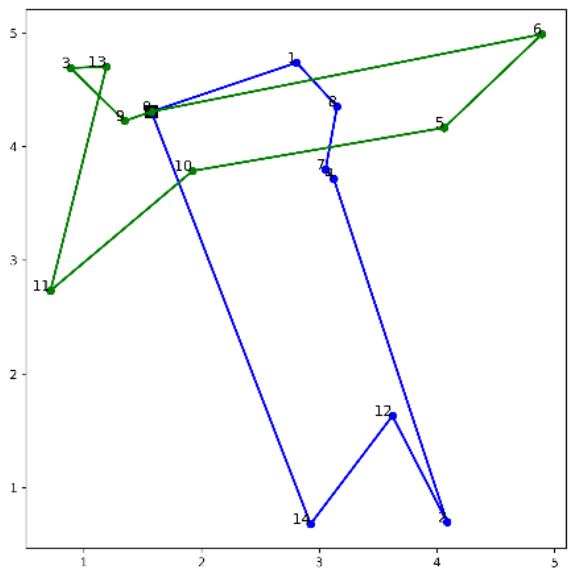

suggested solution #7 score =4

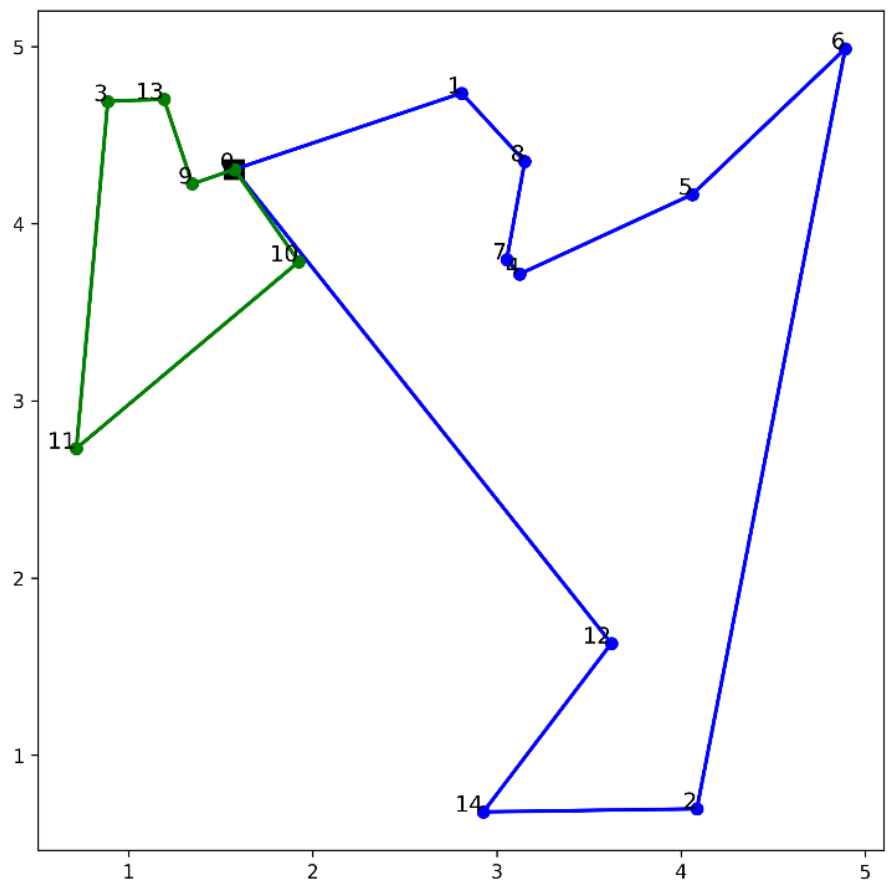

Final solution

**Fig 4.** Visual representations of seven proposed solutions by the critic agent, along with scores assigned by the score agent, highlighting the iterative improvement process in the multi-agent system.

### *4.3 Multi-Agent 1 Example 2*

Figure 6 shows the third iteration of solving a problem with a network comprising 30 nodes, the progression begins with the leading solution from the previous iteration, showcased at the top of the visual table. This serves as the benchmark for the subsequent evaluations conducted by the critic agent. The additional solutions proposed are visually represented and scored based on their coverage and accuracy in addressing the node network. These solutions are listed as follows: Suggested Solution #1 with a score of 3, Suggested Solution #2 with a score of 4, Suggested Solution #3 also scoring 3, Suggested Solution #4 with the lowest score of 1, Suggested Solution #5 scoring 2, Suggested Solution #6 with the highest score of 5, and Suggested Solution #7 scoring 3. The scoring outcomes indicate a clear correlation between the completeness of node coverage and the assigned scores, with missing nodes notably reducing the effectiveness of the solutions. The highest score, a 5, reflects a solution that significantly improves on node coverage, suggesting near-optimal completeness. This methodical, iterative approach, facilitated by the interactions between the critic and scoring agents, underscores the efficacy of the multi-agent system in refining solutions to ensure comprehensive node coverage.

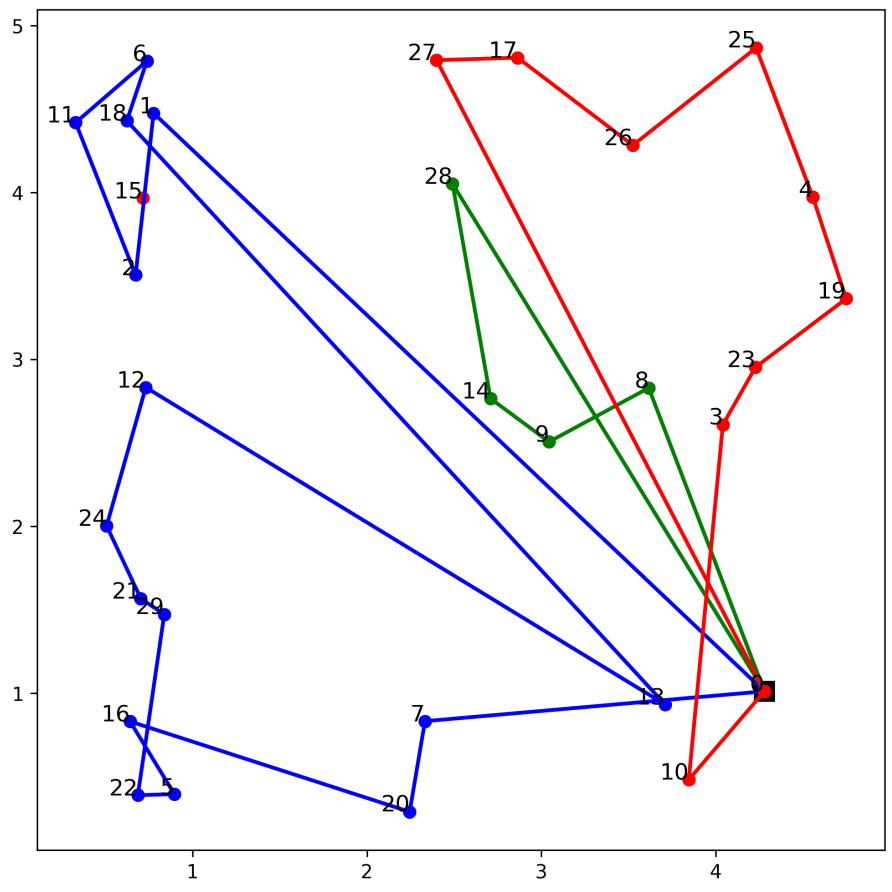

Initial solution

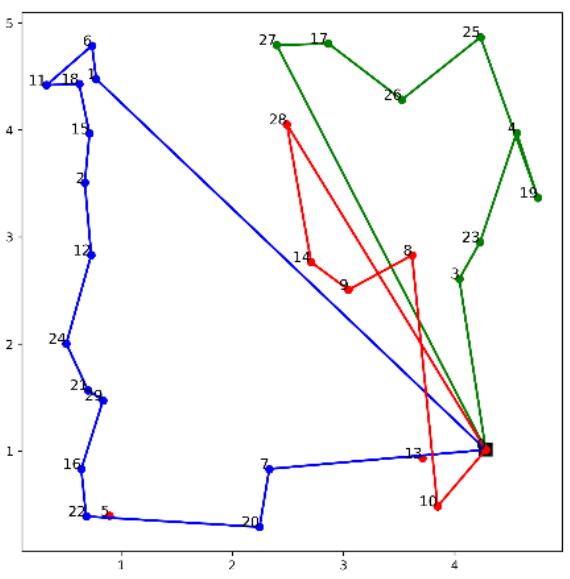

Suggested solution #1 score =3

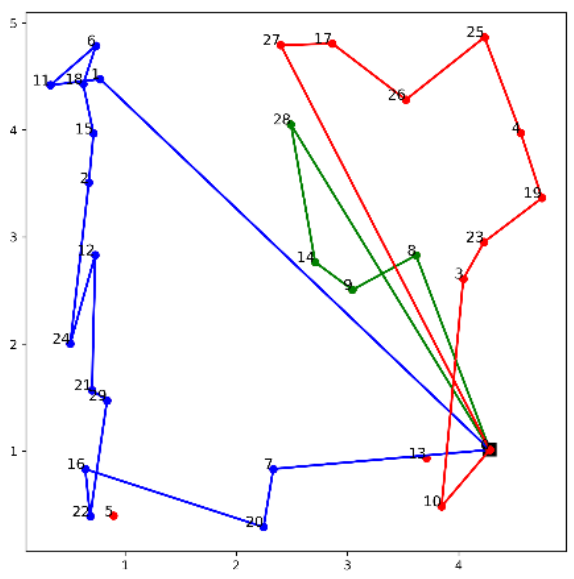

Suggested solution #2 score =4

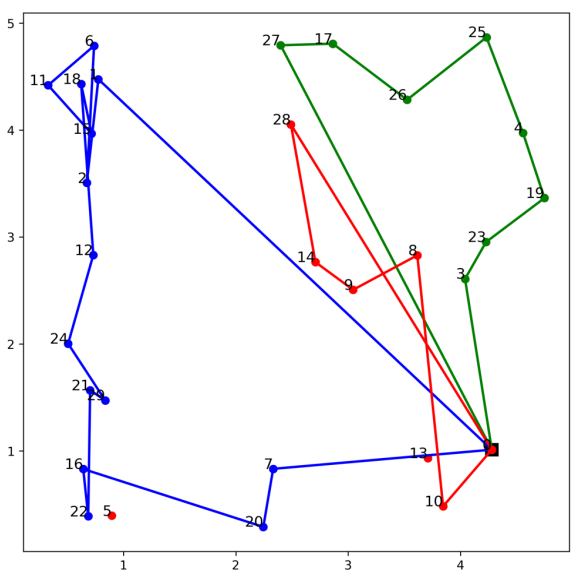

Suggested solution #3 score =3

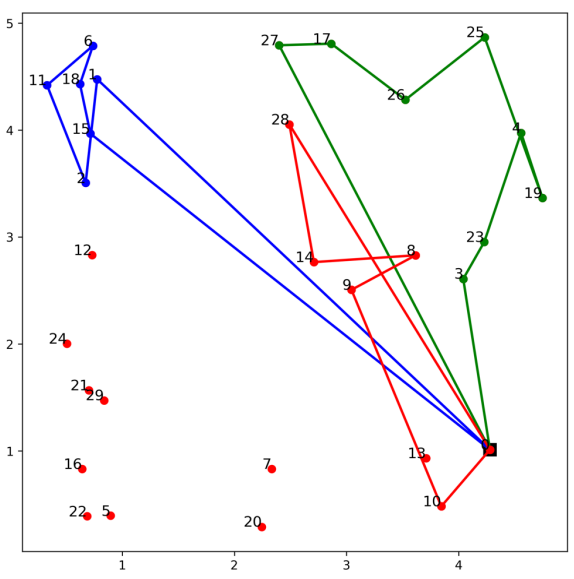

Suggested solution #4 score =1

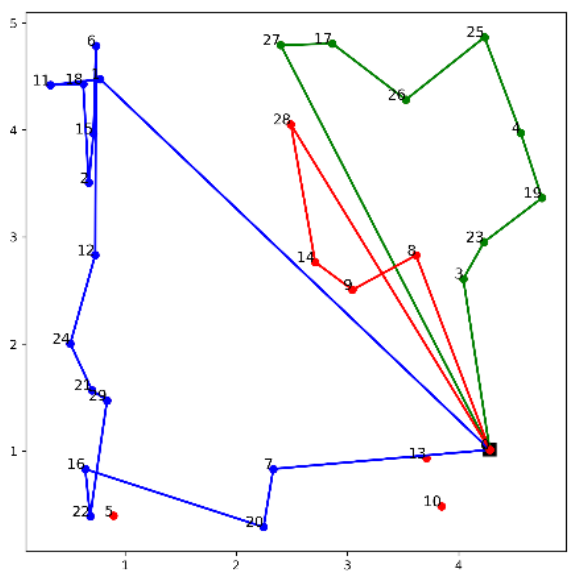

Suggested solution #5 score =2

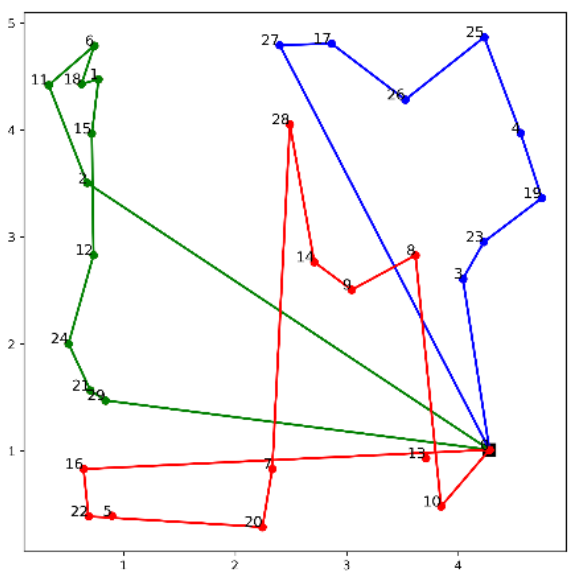

Suggested solution #6 score =5

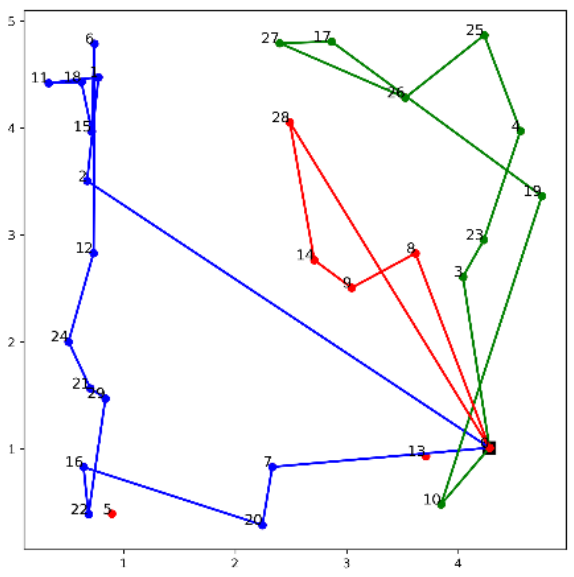

Suggested solution #7 score =3

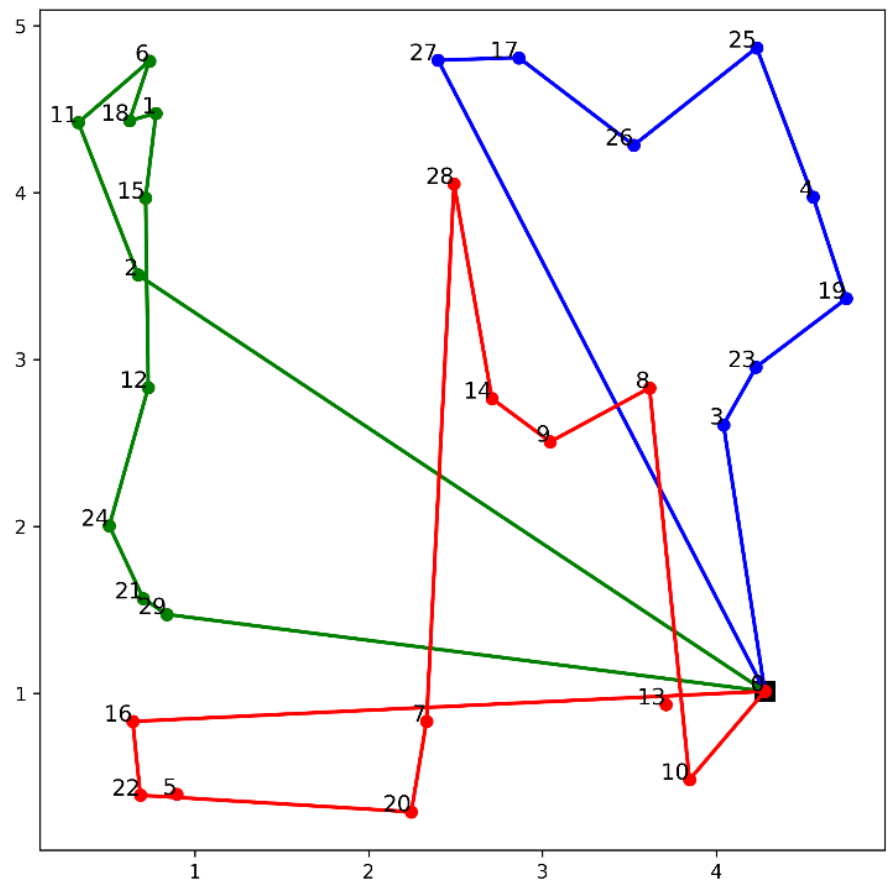

Final solution

**Fig 5.** Evaluation of Proposed Solutions in the Second Iteration with Scores Indicating the Completeness of Node Coverage in a 30-Node Network.

### *3.3 Multi-Agent 2 Solutions Quality*

The analysis of Multi-Agent 2 strategy performance in addressing the m-Salesmen problem demonstrates a significant enhancement in solution quality, particularly when contrasting against the Zero-Shot strategy. As outlined in Fig. 6, the mean gap of gaps across different salesman configurations show that Multi-Agent 2 consistently reduces the solution gaps, even as problem sizes increase. Additionally, the standard deviation of the gap is equal to or improved in Multi-agent 2 compared to the zero-shot strategy. This is indicative of its robustness in managing complex scenarios more effectively than the baseline Zero-Shot approach.

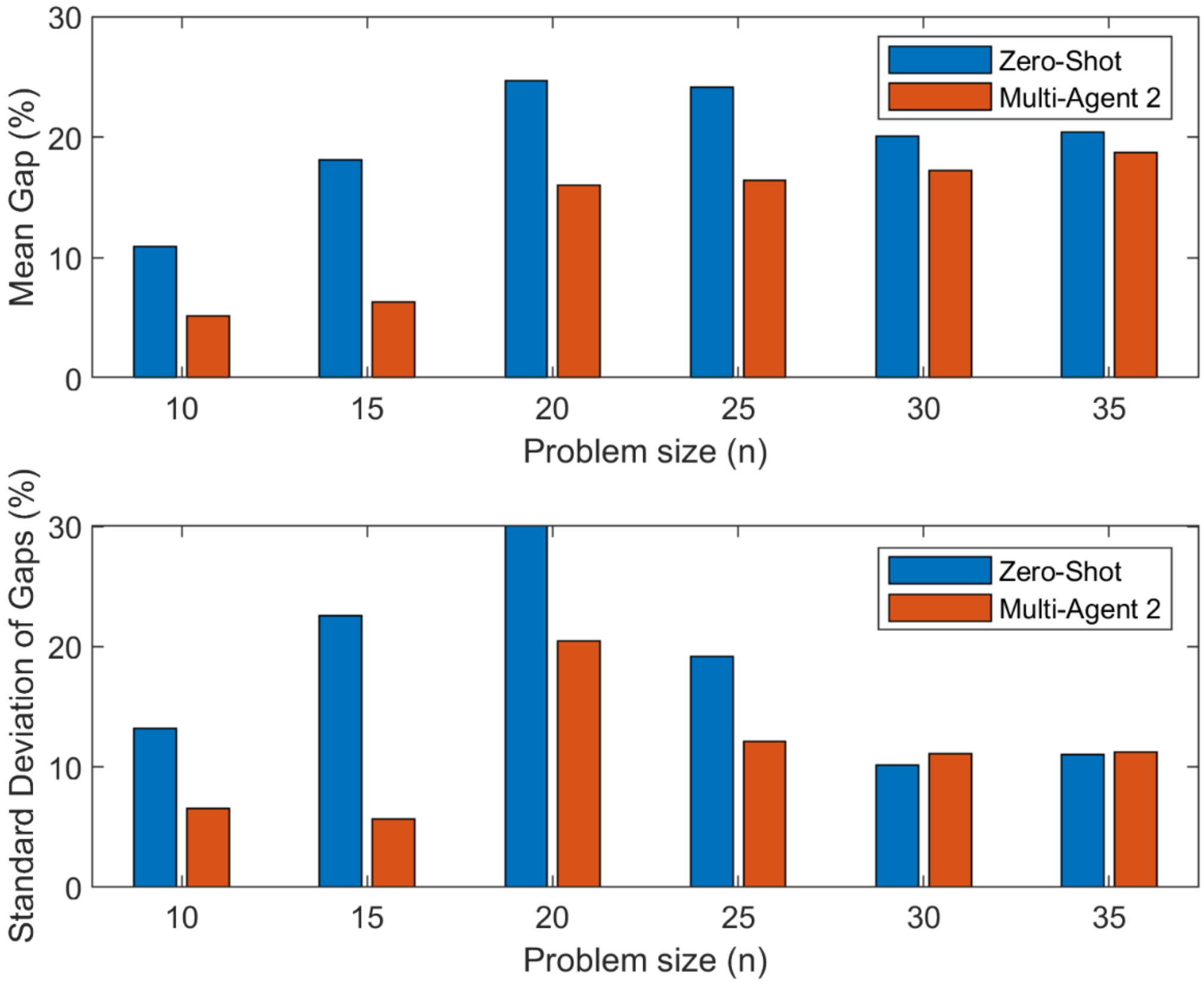

(a)

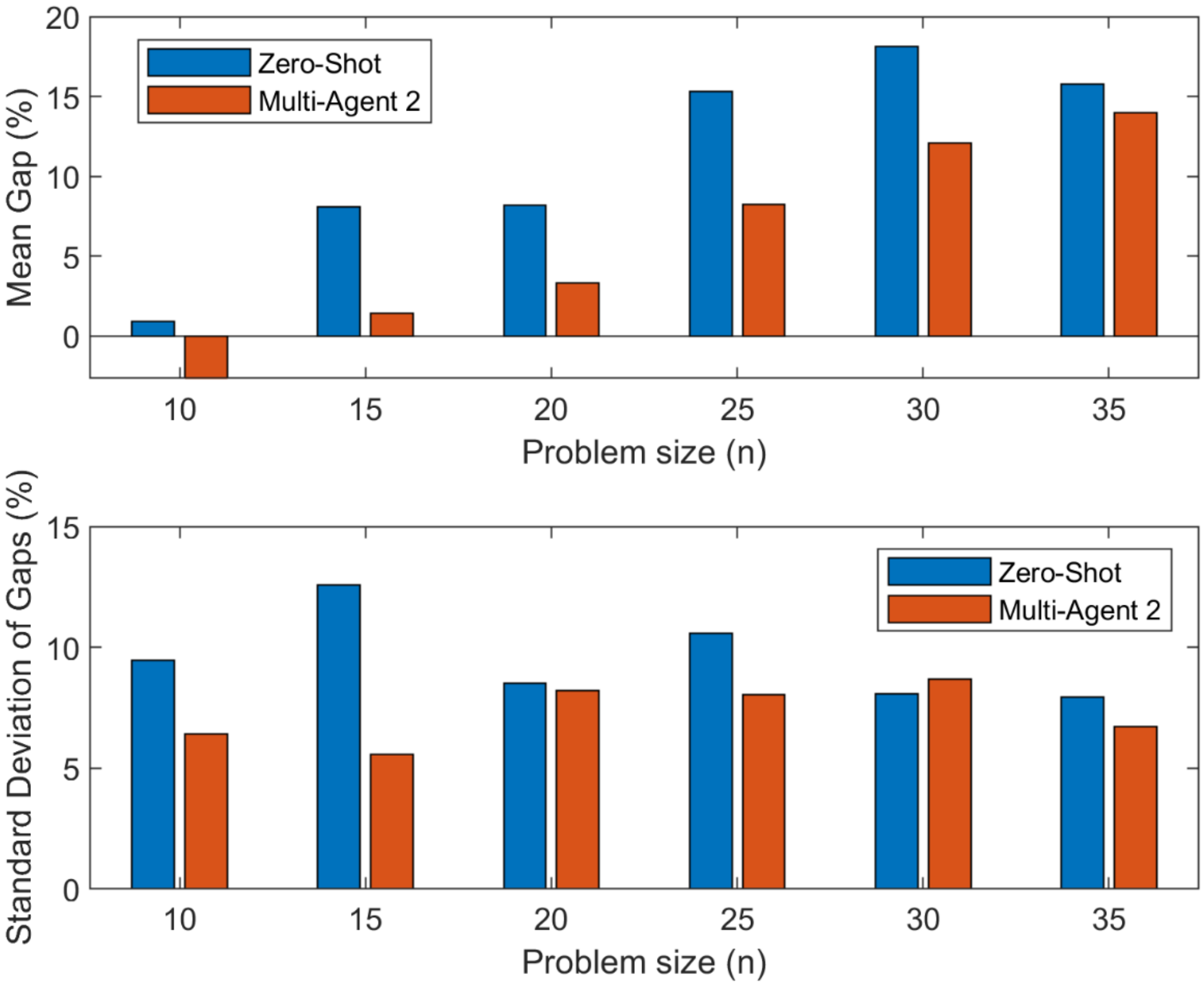

(b)

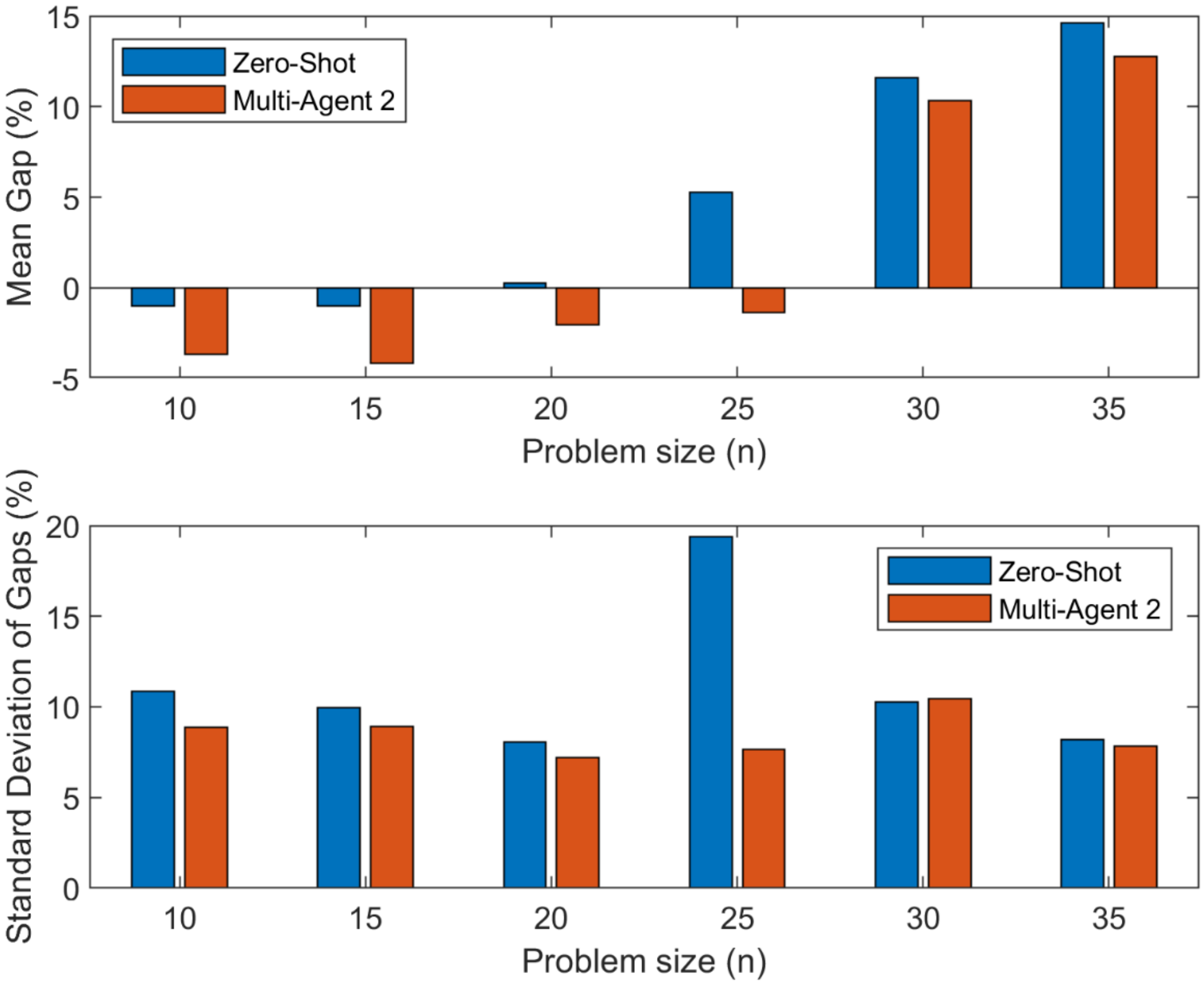

(c)

**Fig 6.** Comparative Performance of Zero-Shot and Multi-Agent Strategy 2 across Different Salesmen Configurations: Analysis of Mean Gap and Standard Deviation by Problem Size (a) one salesman (b) two salesmen (c) three salesmen

The statistical analysis presented in Table 3 further underscores this improvement. Utilizing the Wilcoxon signed-rank test, the results reveal that the differences in route quality between Multi-Agent 2 and Zero-Shot strategies are statistically significant, with p-values consistently lower across various configurations and problem sizes. Notably, the number of pairs remains high across all problem sizes for Multi-Agent 2, suggesting fewer instances of hallucinations by the initializer, which are commonly observed in larger problems. This indicates that Multi-Agent 2, even without the scoring phase by a Scorer Agent, maintains high fidelity in route generation and optimization, emphasizing its efficiency in iterative refinement through the Critic Agent's interventions.

This reduction in hallucinations and the subsequent improvement in route quality highlight Multi-Agent 2's effectiveness in leveraging visual reasoning for complex route optimization tasks. The streamlined process, focusing solely on the Initializer and Critic Agents, proves to be highly effective, particularly in environments where rapid route generation and iterative refinement are crucial. These findings advocate for the potential of simplified visual reasoning models in solving not just theoretical problems but potentially real-world logistical challenges.

**Table 3.** Statistical Analysis of Route Quality Improvement Using Multi-Agent 2 Strategy Over Zero-Shot

| | **m=1** | | **m=2** | | **m=3** | |
|---|---|---|---|---|---|---|
| problem size | p-value | number of pairs | p-value | number of pairs | p-value | number of pairs |
| 10 | 0.0004 | 30 | 0.0001 | 30 | $<0.0001$ | 30 |
| 15 | 0.0001 | 28 | 0.0002 | 26 | 0.0001 | 25 |
| 20 | 0.0001 | 24 | 0.0005 | 20 | 0.0039 | 19 |
| 25 | 0.0020 | 19 | 0.0002 | 19 | 0.0005 | 19 |
| 30 | 0.0020 | 15 | 0.0078 | 16 | 0.1250 | 21 |
| 35 | 0.0313 | 17 | 0.1250 | 16 | 0.0625 | 14 |

Fig8 illustrates the difference in the mean gap percentage between the zero-shot strategy and the two multi-agent strategies across varying problem sizes for $m = 1$, $m = 2$, and $m = 3$ salesmen. The analysis focuses on comparing the improvement each multi-agent strategy offers over the zero-shot baseline by evaluating the reduction in the mean gap percentage.

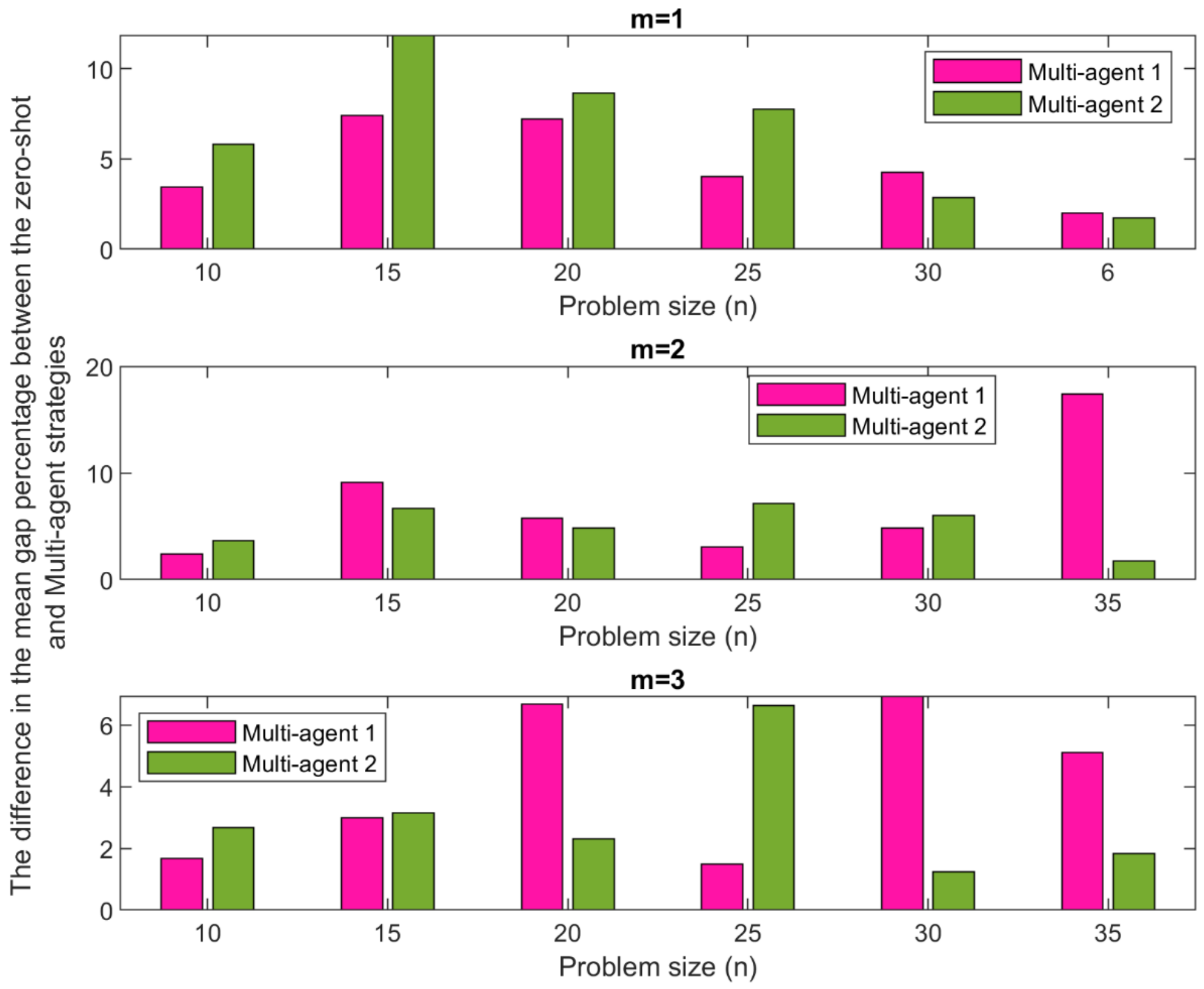

**Fig 7.** Comparative Analysis of Mean Gap Percentage Reductions: Evaluating Multi-Agent 1 and Multi-Agent 2 Against Zero-Shot Across Different Problem Sizes and Salesman Configurations

For $m = 1$ (i.e., single salesman), Multi-Agent 2 generally outperforms Multi-Agent 1 at smaller problem sizes, indicating a more effective optimization in simpler scenarios. However, as the problem size increases, the performance advantage of Multi-Agent 2 diminishes slightly, suggesting that Multi-Agent 2 strategy may be better suited for less complex problems. As the number of salesmen increases to two and three (i.e., $m = 2$ and $m = 3$), the performance dynamics shift. Multi-Agent 1 starts to show more significant improvements over the zero-shot, particularly in medium-sized problems. For instance, at a problem size of 35 for $m = 2$, Multi-Agent 1 substantially outperforms Multi-Agent 2, suggesting that its strategies may be better suited to handling more complex scenarios with multiple salesmen. This visualization underscores that while Multi-Agent 2 excels in simpler setups, Multi-Agent 1 may offer more robust solutions in more challenging environments, effectively addressing the zero-shot's shortcomings in larger and more complex problem configurations

### *4.4 Multi-Agent 2 Example 1*

Fig. 8 illustrates the Multi-Agent 2 solution using a complex network comprising 30 nodes. This figure showcases the ongoing refinement, beginning with the leading solution from the previous iteration, which serves as a benchmark for subsequent evaluations.

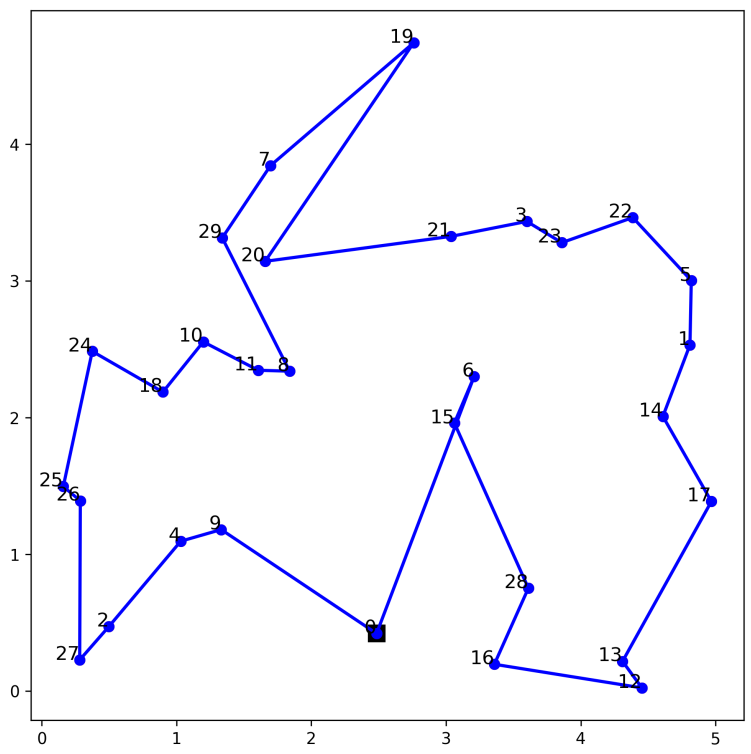

Initializer solution

$(total\ distance\ = 24.24)$

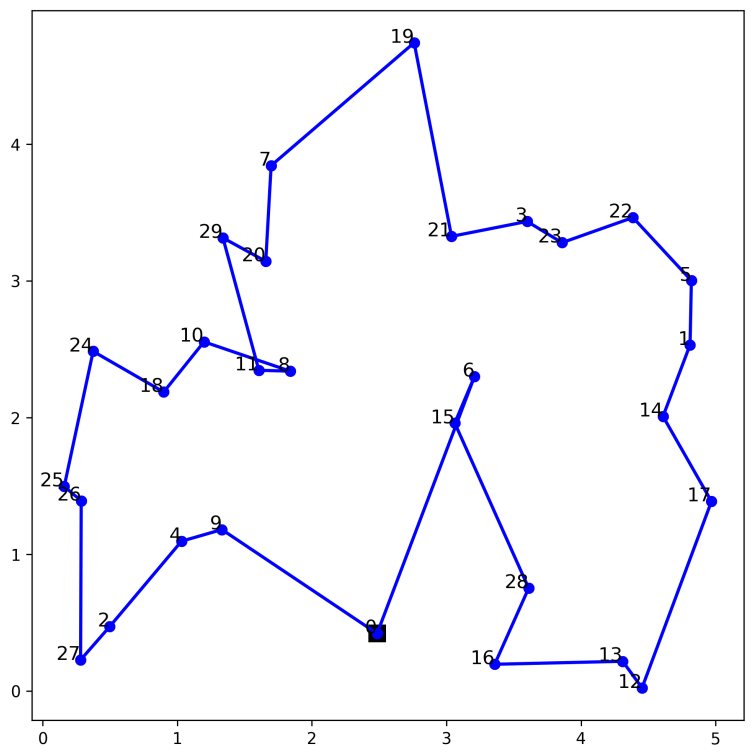

Critic solution iteration #1

$(total\ distance = 22.86)$

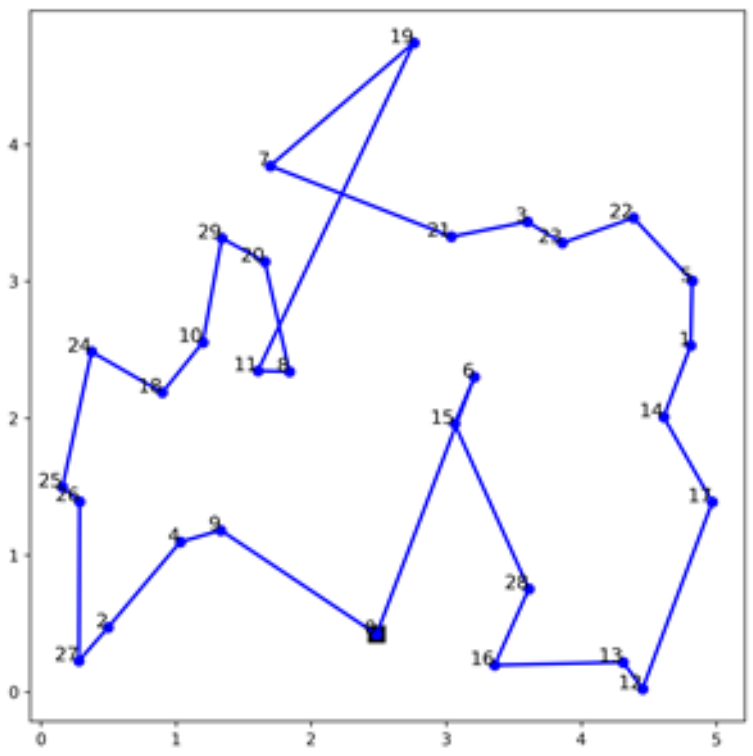

Critic solution iteration #2

$(total\ distance = 24.73)$

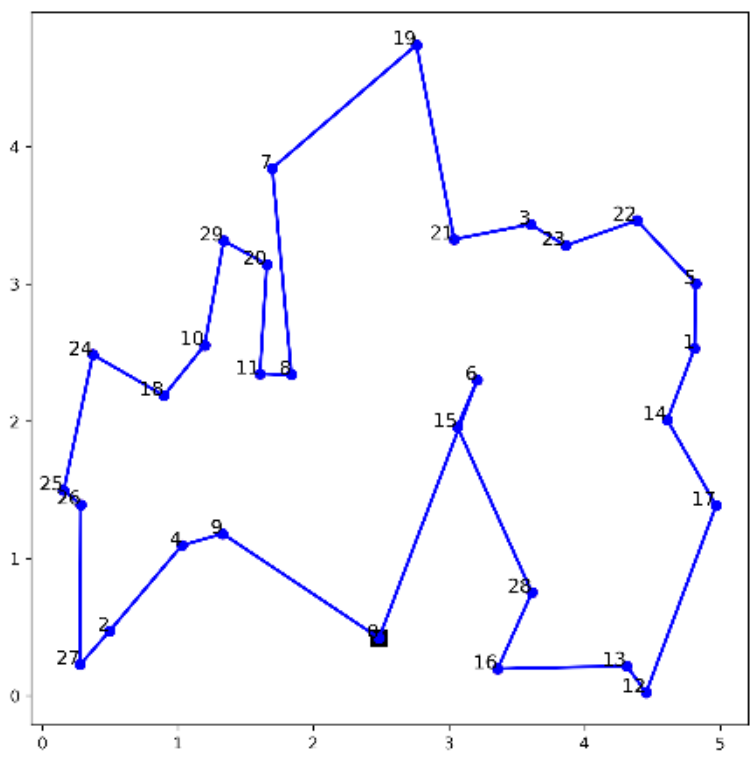

Critic solution iteration #3

$(total\ distance = 23.56)$

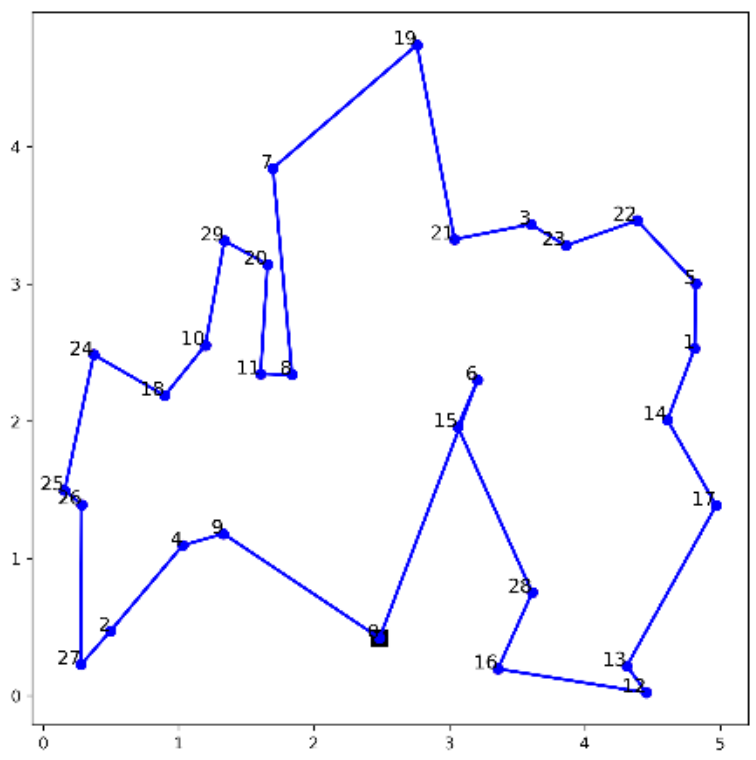

Critic solution iteration #4

$(total\ distance = 23.60)$

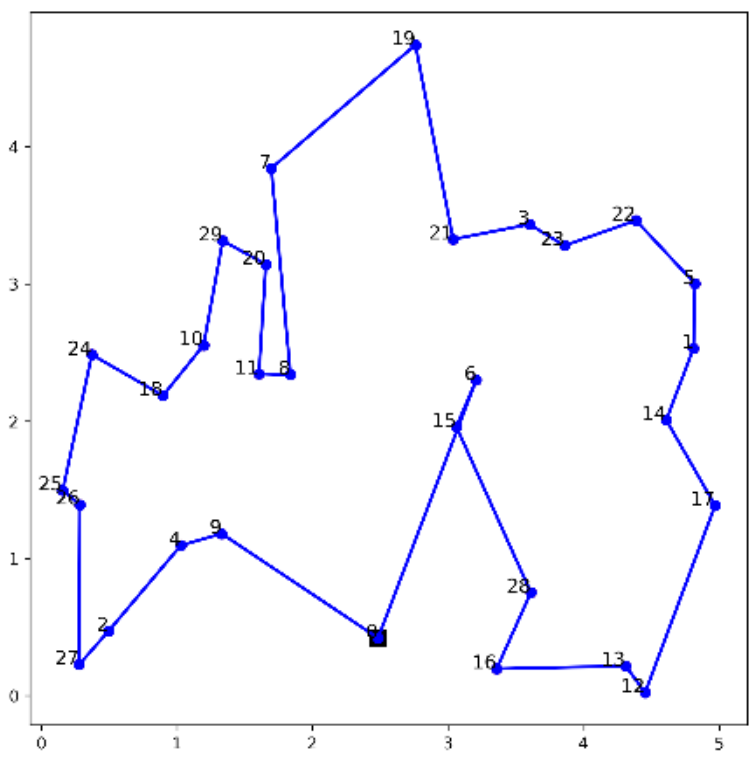

Critic solution iteration #5

$(total\ distance = 23.56)$

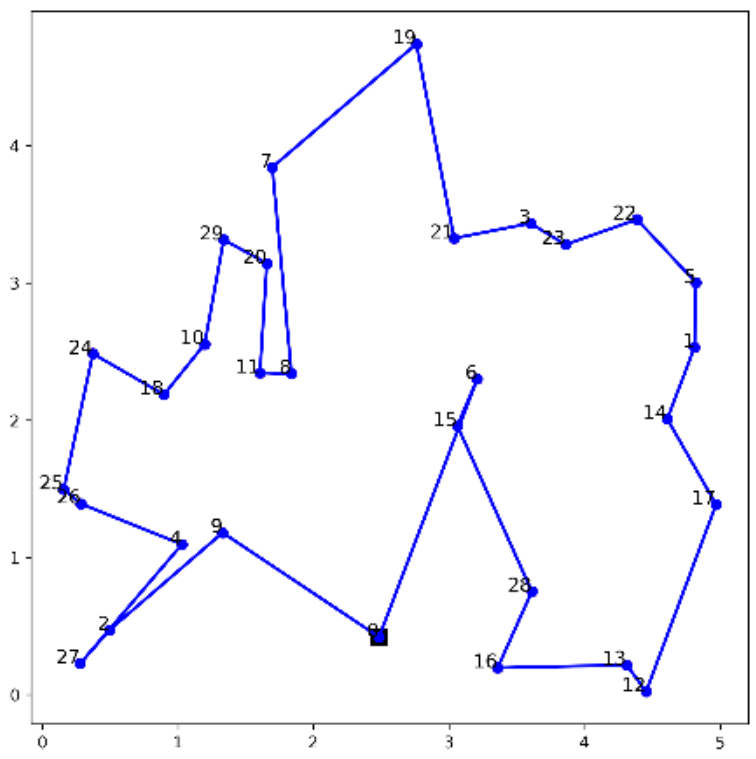

Critic solution iteration #6

$(total\ distance = 24.31)$

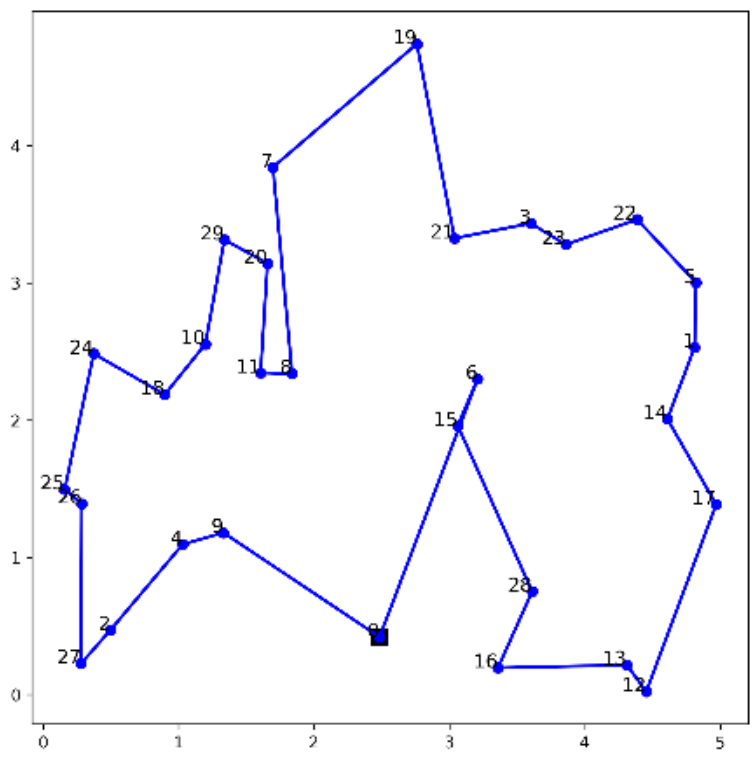

Critic solution iteration #7

$(total\ distance = 23.56)$

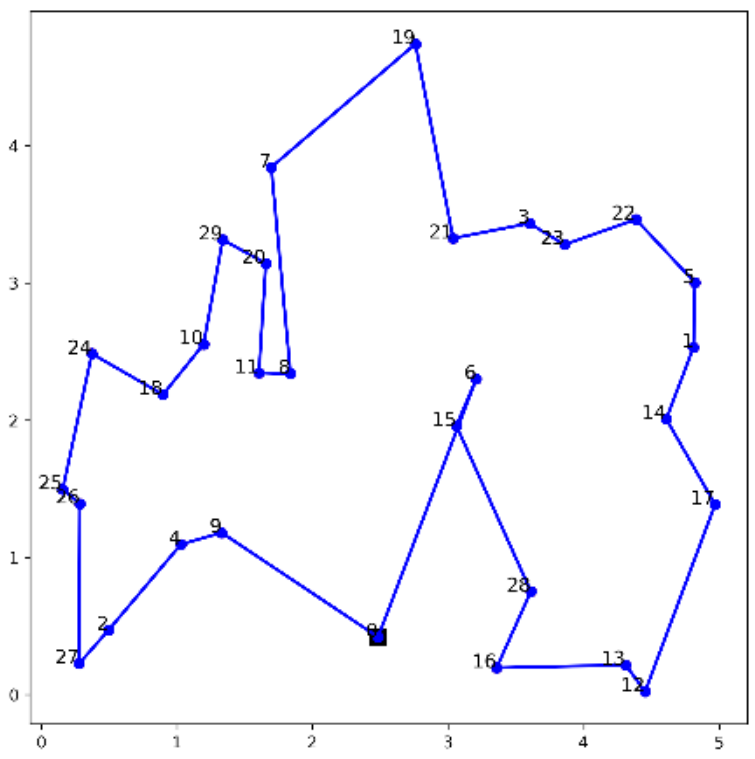

Critic solution iteration #8

$(total\ distance = 23.56)$

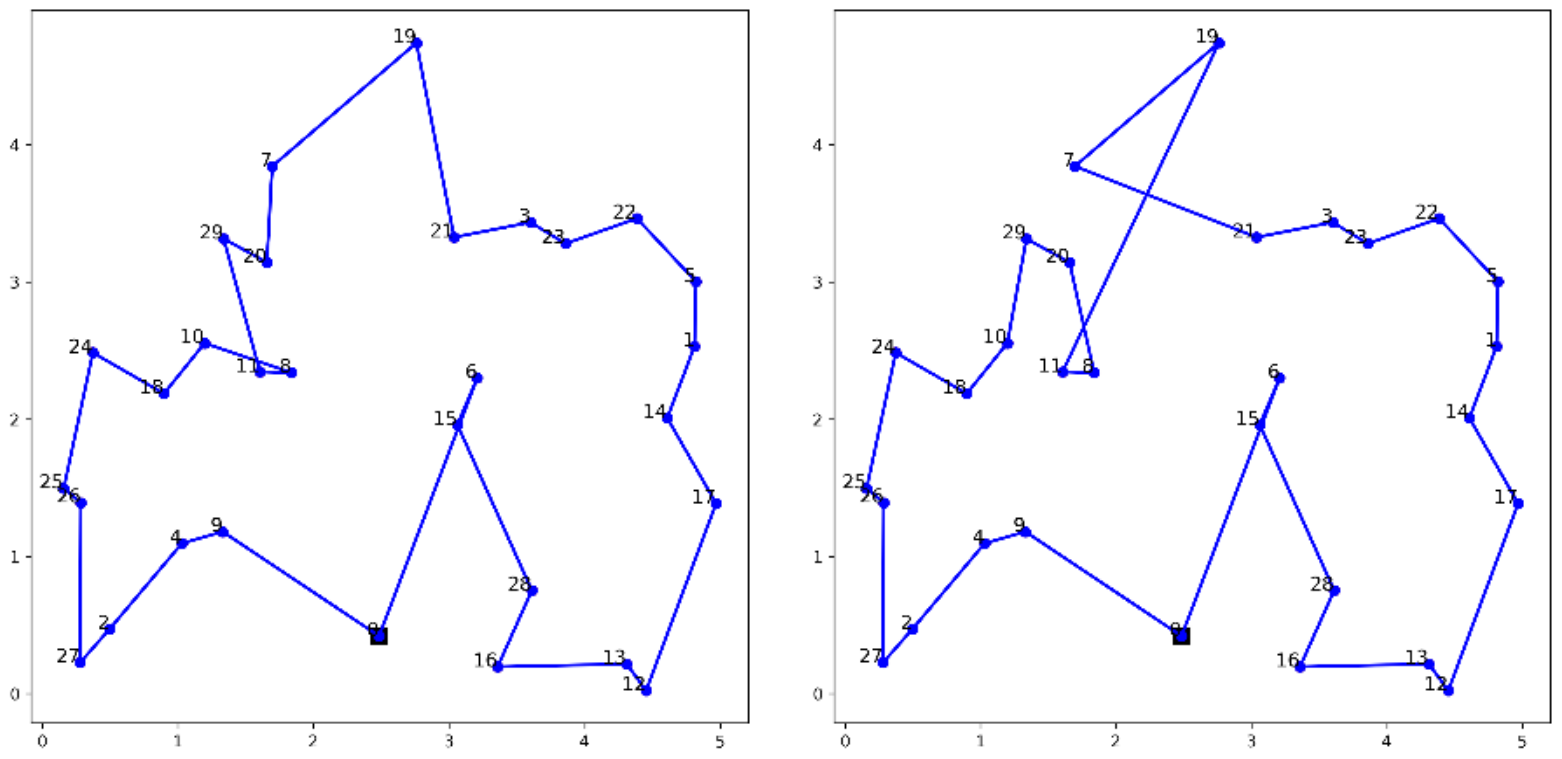

Critic solution iteration #9 ($total\ distance = 22.86$)

Critic solution iteration #10 ($total\ distance = 24.73$)

**Fig 8.** Comparative Analysis of Mean Gap Percentage Reductions: Evaluating Multi-agent 1 and Multi-agent 2 Against Zero-Shot Across Different Problem Sizes and Salesman Configurations

The Multi-Agent 2 strategy, which involves only the initializer and critic agents, has been shown to be effective in refining routes for a given problem in the previous example. To better understand how this strategy works, let's examine the example in detail. By employing an iterative and systematic approach, the dynamic interaction between the critic and scoring agents effectively highlights the efficacy of the multi-agent system in refining and optimizing solutions. This method ensures comprehensive node coverage and optimizes network connectivity, which is essential for complex system analyses in network optimization studies. In this strategy, the initializer starts with a solution that has a distance of 24.24 units. Over the course of ten iterations, the critic agent works to refine the initial path, and the first iteration achieves the most significant reduction in distance, bringing it down to 22.86 units, which is the optimal outcome in the series of iterations. However, subsequent iterations display inconsistent results, with distances fluctuating and even reverting to 24.73 units at times. This variability underscores the potential of the critic agent to enhance the initial route proposed by the initializer, but it also highlights the inconsistency in the optimization process across different iterations.

### *4.5 Multi-Agent 2 Example 2*

Fig. 9 illustrates the intricate and evolving process of route optimization through the Multi-Agent 2 strategy. This example highlights the critic agent's exceptional ability to correct initial errors and make necessary adjustments. The series of iterations exemplifies the critic agent's thorough and deliberate approach to identifying and correcting misconceptions, particularly in instances where vital junctions were omitted during the initial proposals.

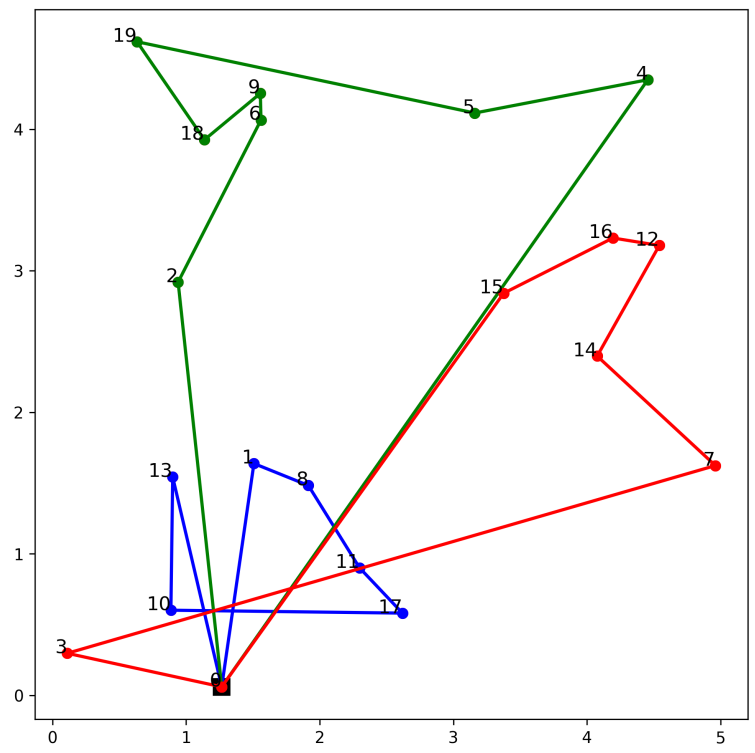

Initializer solution

$(total\ distance\ = 35.43)$

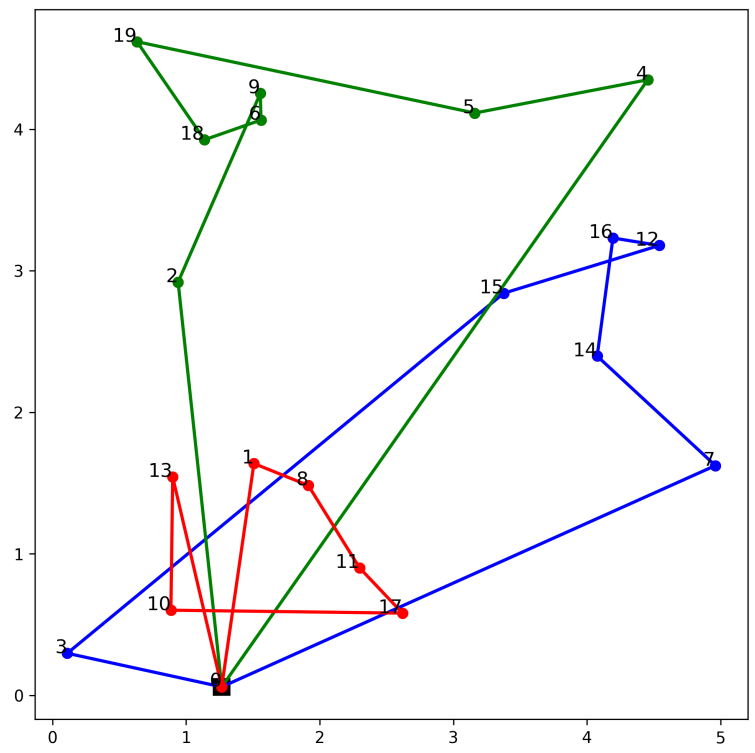

Critic solution iteration #1

$(total\ distance = 35.38)$

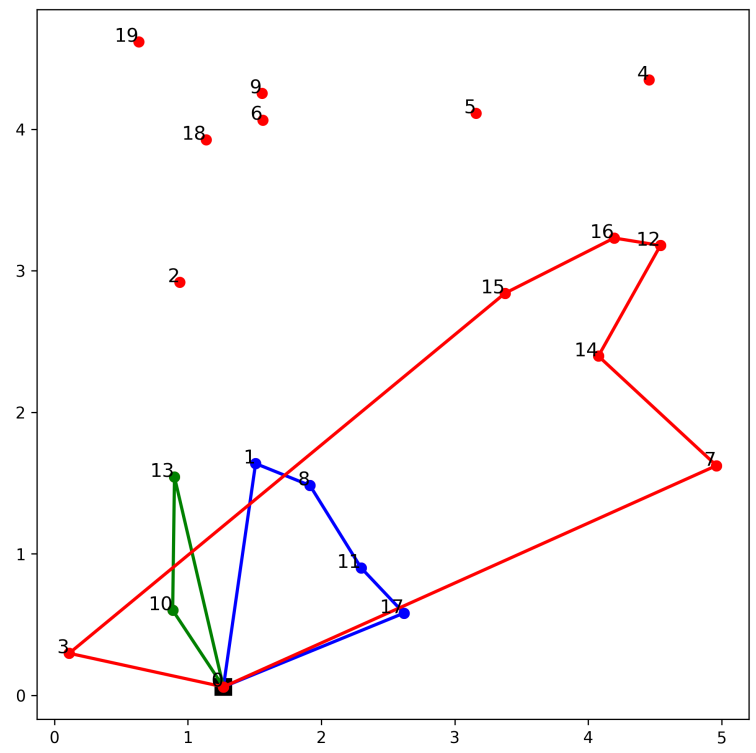

Critic solution iteration #2

$(total\ distance = NaN)$

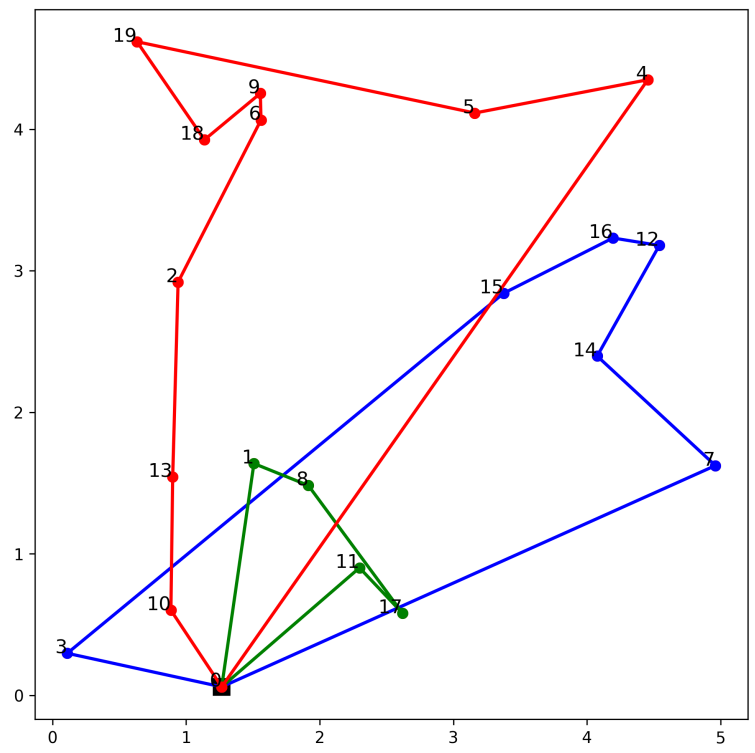

Critic solution iteration #3

$(total\ distance = 32.74)$

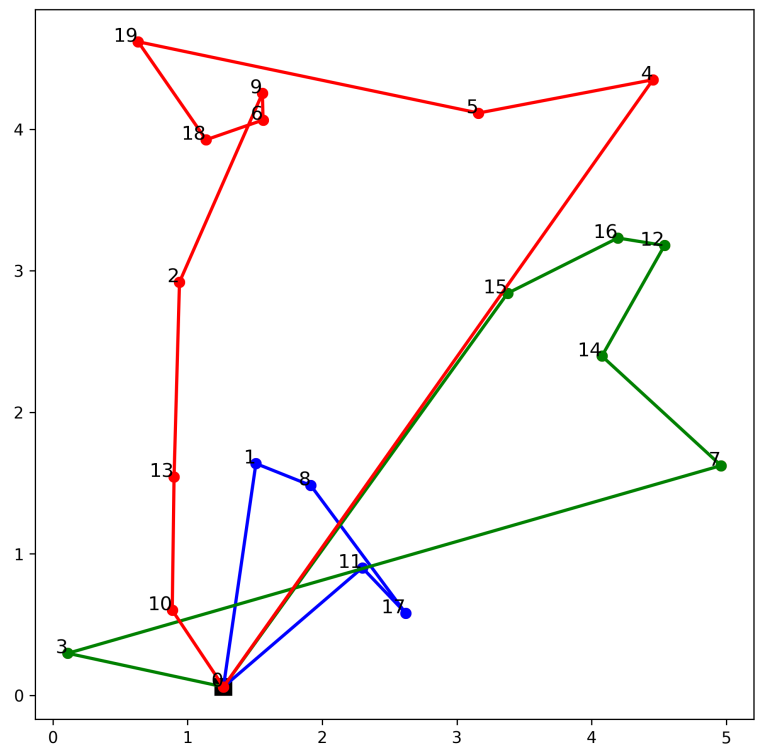

Critic solution iteration #4

$(total\ distance = 33.19)$

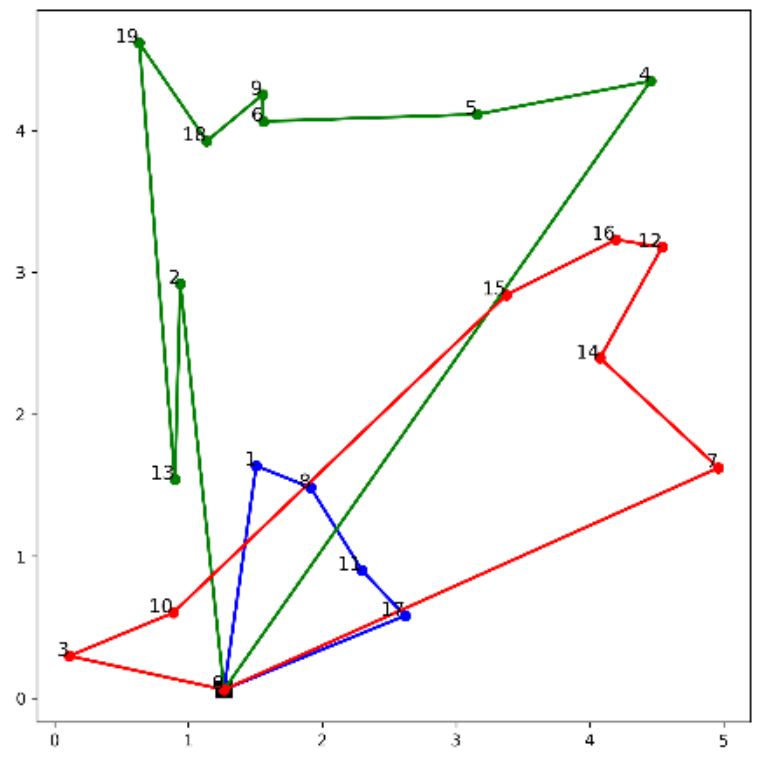

Critic solution iteration #5

$(total\ distance = 34.53)$

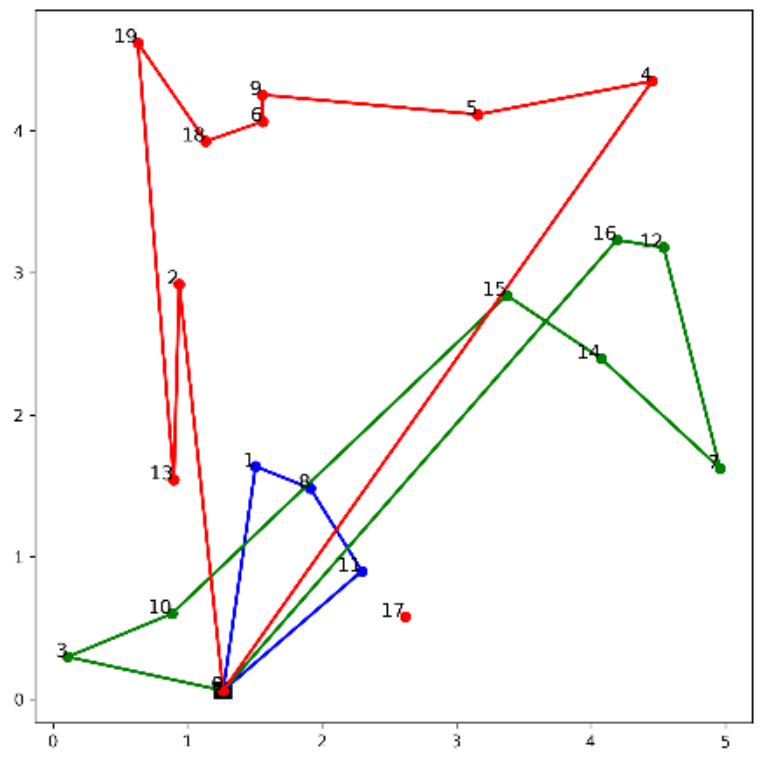

Critic solution iteration #6

$(total\ distance = NaN)$

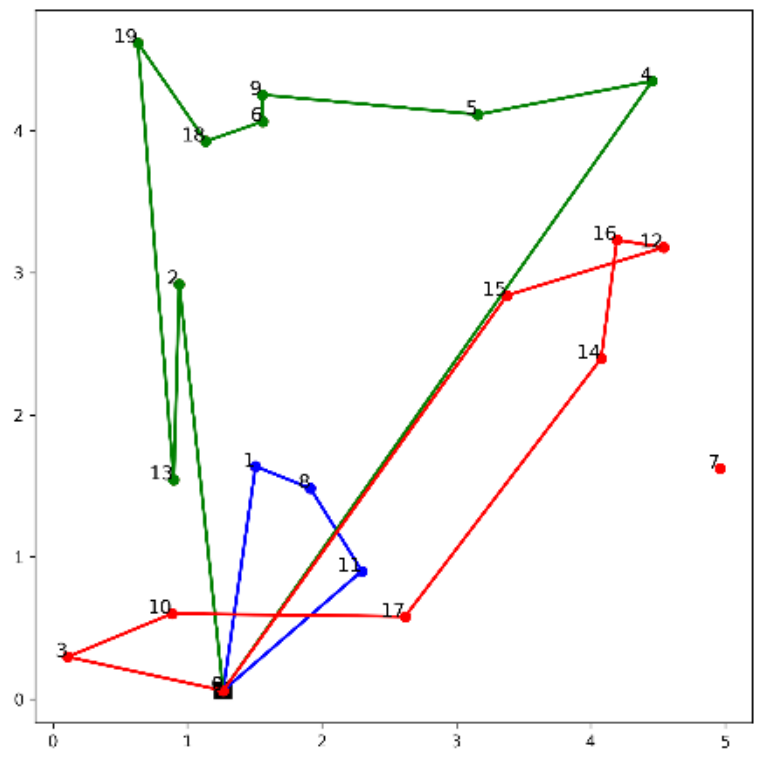

Critic solution iteration #7

$(total\ distance = NaN)$

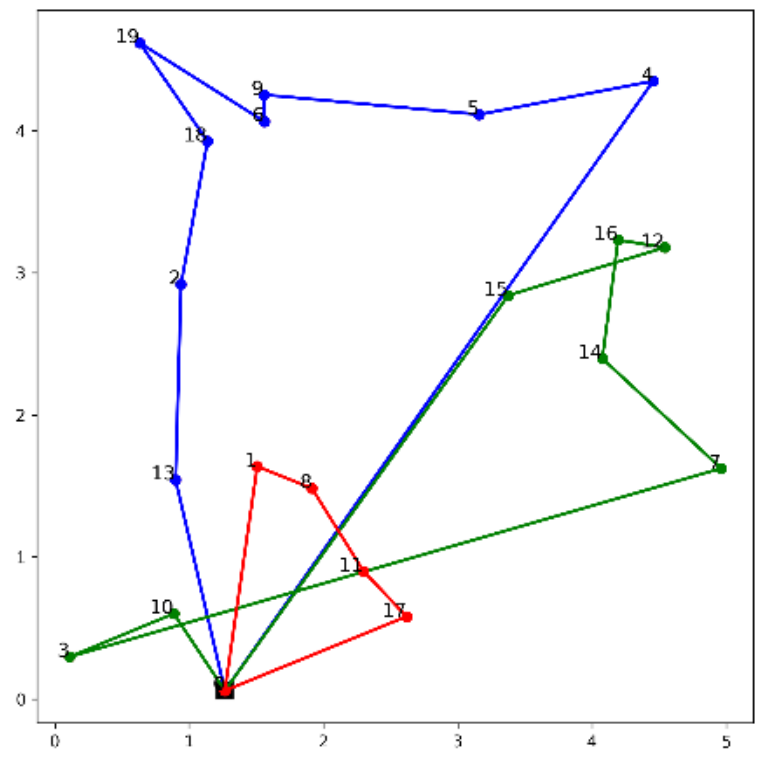

Critic solution iteration #8

$(total\ distance = 32.56)$

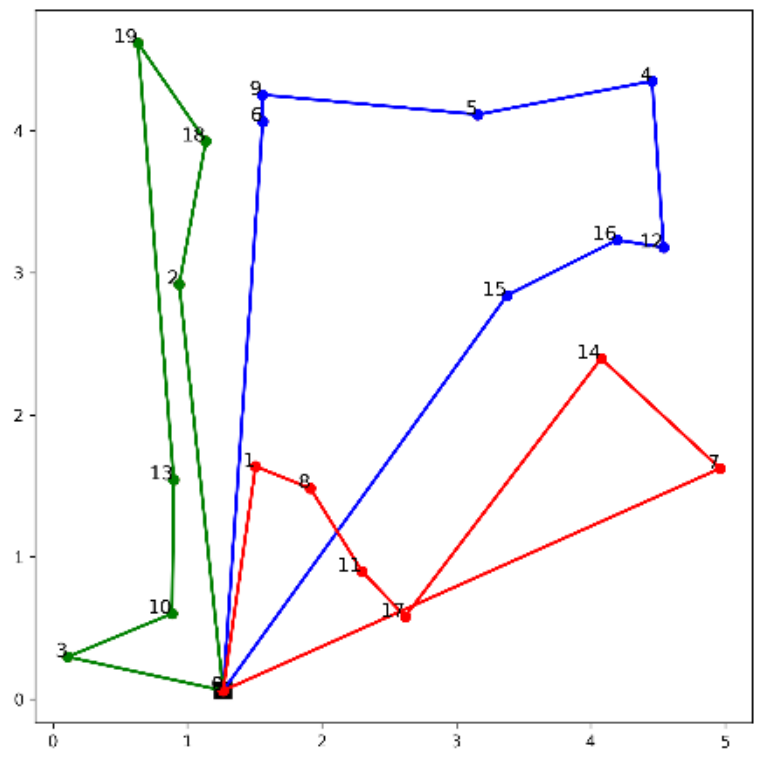

Critic solution iteration #9

$(total\ distance =\ 34.56)$

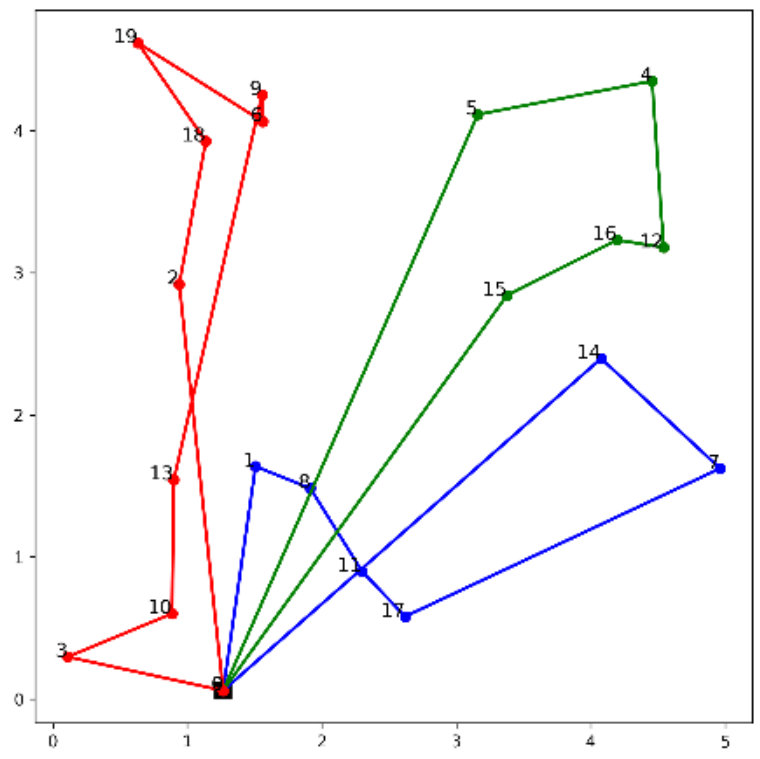

Critic solution iteration #10

$(total\ distance =\ 34.07)$

**Fig 9.** Comparative Analysis of Mean Gap Percentage Reductions: Evaluating Multi-agent 1 and Multi-agent 2 Against Zero-Shot Across Different Problem Sizes and Salesman Configurations.

Initially, the initializer provides a solution with a total distance of 35.43 units. The critic's first two iterations make minor adjustments with negligible reductions in total distance, reflecting the critic's initial struggle to optimize the route effectively. Notably, iterations 2 and 6 demonstrate the occurrence of hallucinations, where the critic agent fails to include all nodes, leading to a $NaN$ (i.e., Not a Number) total distance, indicating an incomplete or erroneous route computation.

Despite these setbacks, the critic agent remarkably recovers in subsequent iterations. Iteration 3 shows a significant improvement with a reduced total distance of 32.74 units, Iterations 4 through 8 oscillate in effectiveness, but each maintains a coherent route. Following other hallucinations in iterations 6 and 7, the agent recovers again in iterations 9 and 10, finalizing with a distance of 34.07 units, demonstrating an overall enhancement from the initial route.

## 5. Conclusion

This study introduces a pioneering framework for tackling the TSP and mTSP solely through visual reasoning using ChatGPT-4o as an example of MLLMs, marking a departure from conventional reliance on numerical data. By leveraging a multi-agent system in MLLMs, our approach systematically improves route quality by iteratively refining proposed solutions. This iterative process ensures comprehensive node coverage, minimal route intersections, and optimized path lengths without explicit computation of distances. Comparative analyses underscore the competitiveness of our approach against industry-standard tools, validating its potential for practical applications in diverse domains requiring efficient routing solutions

Results showed that two multi-agent models including Multi-Agent 1, which contains Initializer, Critic, and Scorer; and Multi-Agent 2, which only contains Initializer and Critic—both led to a significant enhancement in the quality of solutions for TSP and mTSP problems. Using MLLM as an agent utilized visual reasoning to bypass traditional computational complexities, offering a novel problem-solving method that mirrors human intuitive processes. The results indicate that Multi-Agent 1, with its trio of Initializer, Critic, and Scorer agents, excels in environments necessitating strict route refinement and evaluation, suggesting a robust framework for managing intricate optimization situations. On the other hand, Multi-Agent 2 streamlines the optimization process by concentrating on iterative refinements conducted by the Initializer and Critic, proving effective in fast decision-making contexts.

Nevertheless, as problem sizes increase, the emergence of hallucinations—instances where routes fail to visit all designated nodes—highlights a critical flaw in the scaling capabilities of the proposed strategies using MLLMs. This phenomenon raises questions about the dependability of these systems in larger, more intricate situations, a significant limitation in real-world applications like urban planning and logistics. Future research includes exploring other open-source MLLMs. Another promising avenue is experimenting our methodology on 3-dimensional data points, as well as using other valuation metrics such as the combination of time and distance cost.

**Author Contributions:** Conceptualization, M. E., A. A., T. I. A., A. J., H. I. A., S. J., A. A., S. G., and A. R.; methodology, M. E.; software, M. E.; validation, M. E., A. A., T. I. A., A. J., H. I. A., and S. J.; resources, M. E., S. G., and A. R.; writing—original draft preparation, A. A., T. I. A., A. J., H. I. A., S. J., and A. A.; writing—review and editing, H. I. A., S. G., and A. R.; visualization, A. A.; supervision, S. G., and A. R.; project administration, S. G., and A. R.; funding acquisition, M. E., S. G., and A. R. All authors have read and agreed to the published version of the manuscript.

**Funding:** This research received no external funding.

**Data Availability Statement:** Data is available upon request.

**Acknowledgments:**

**Conflicts of Interest:** The authors declare no conflicts of interest.